\documentclass[10pt, a4paper]{article}
\usepackage{placeins} 
\usepackage{comment}
\usepackage{float}
\usepackage{tabularx}
\usepackage{tcolorbox}
\tcbuselibrary{breakable, skins}
\usepackage{makecell}
\usepackage{amsmath}
\usepackage{enumitem}
\usepackage{booktabs}
\usepackage{multirow}
\usepackage{fontawesome}
\usepackage{array}
\usepackage{tikz}
\newcommand{\sgrades}{S-GRADES} 
\usepackage{pgfplots}
\pgfplotsset{compat=1.18}

\usetikzlibrary{shapes.geometric, arrows.meta, positioning}

\tikzset{
    stage/.style={
        rectangle, rounded corners, draw=black, fill=gray!10,
        text width=10cm, minimum height=1.2cm, align=left, font=\small
    },
    arrow/.style={thick, -{Latex[length=3mm]}}
}

\usepackage{latex/lrec2026} 

\newcommand{\githubicon}{\raisebox{-0.2\height}{\includegraphics[height=1em]{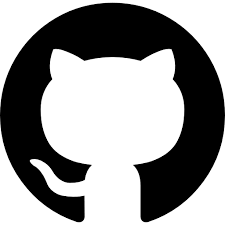}}}

\title{S-GRADES -- Studying Generalization of Student Response Assessments in Diverse Evaluative Settings}

\name{Tasfia Seuti, Sagnik Ray Choudhury} 
\address{
University of North Texas, Denton, TX, USA \\
\texttt{TasfiaSeuti@my.unt.edu, Sagnik.Raychoudhury@unt.edu}
}

\abstract{
Evaluating student responses, from long essays to short factual answers, is a key challenge in educational NLP. Automated Essay Scoring (AES) focuses on holistic writing qualities such as coherence and argumentation, while Automatic Short Answer Grading (ASAG) emphasizes factual correctness and conceptual understanding. Despite their shared goal, these paradigms have progressed in isolation with fragmented datasets, inconsistent metrics, and separate communities. We introduce \sgrades\ (Studying Generalization of Student Response Assessments in Diverse Evaluative Settings), a web-based benchmark that consolidates 14 diverse grading datasets under a unified interface with standardized access and reproducible evaluation protocols. The benchmark is fully open-source and designed for extensibility, enabling continuous integration of new datasets and evaluation settings. To demonstrate the utility of \sgrades, we evaluate three state-of-the-art large language models across the benchmark using multiple reasoning strategies in prompting. We further examine the effects of exemplar selection and cross-dataset exemplar transfer. Our analyses illustrate how benchmark-driven evaluation reveals reliability and generalization gaps across essay and short-answer grading tasks, highlighting the importance of standardized, cross-paradigm assessment. \\
}
\begin{document}

\maketitleabstract

\section{Introduction}
Summative evaluation of student responses encompasses a diverse spectrum of assessment types, from extended essays demonstrating argumentative and analytical skills to concise short-answer responses testing factual knowledge and conceptual understanding. This diversity poses a central challenge in educational Natural Language Processing (NLP), as different response types demand distinct evaluation criteria. Research in this area has developed along two complementary trajectories. \textbf{Automated Essay Scoring (AES)} focuses on evaluating extended writing for coherence, organization, and argument quality, while \textbf{Automatic Short Answer Grading (ASAG)} targets concise factual responses that require concept-level verification rather than stylistic assessment \citep{burrows2015eras}. ASAG tasks also include domain-specific settings such as conceptual assessment of chemistry \citep{sonkar2024automated}, physics reasoning \citep{kortemeyer2024performance}, and conceptual understanding of computer science \citep{xie2024grade}. Despite their shared goal of automating human evaluation for long-form (non-MCQ) responses, AES and ASAG have largely progressed in isolation, supported by separate datasets, metrics, and research communities (\S \ref{sec:related_work}). No previous work has unified diverse assessment tasks under a single blind-test evaluation.

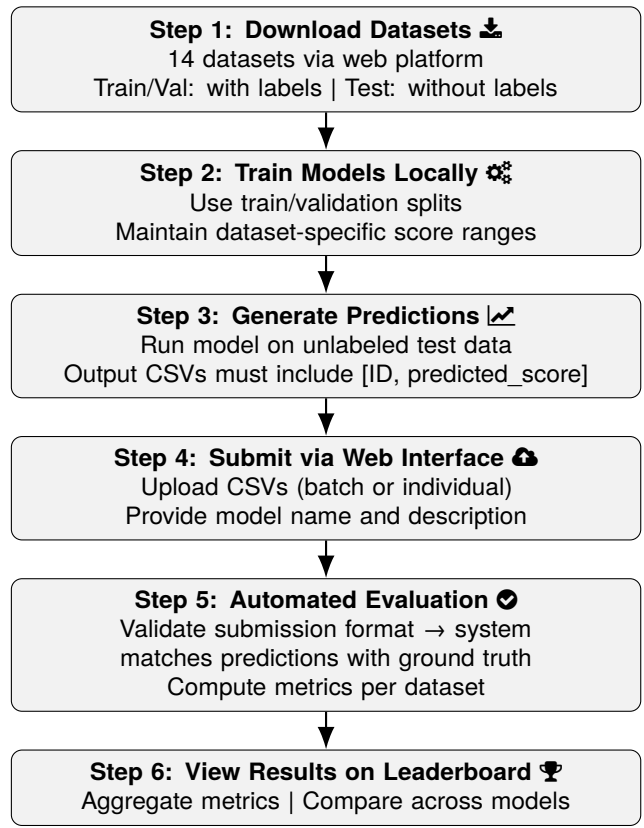
\begin{figure}[!htbp]
\centering
\footnotesize
\begin{tikzpicture}[scale=0.7, node distance=0.5cm]
\tikzset{
    stage/.style={
        rectangle, rounded corners, draw=black, fill=gray!10,
        text width=8cm,  
        minimum height=1cm,
        align=center,    
        font=\small,
        inner sep=4pt
    },
    arrow/.style={thick, -{Latex[length=3mm]}}
}

\node[stage] (step1) {\textbf{Step 1: Download Datasets \faDownload}\\14 datasets via web platform\\Train/Val: with labels | Test: without labels};
\node[stage, below=of step1] (step2) {\textbf{Step 2: Train Models Locally \faCogs}\\Use train/validation splits\\Maintain dataset-specific score ranges};
\node[stage, below=of step2] (step3) {\textbf{Step 3: Generate Predictions \faLineChart}\\Run model on unlabeled test data\\Output CSVs must include [ID, predicted\_score]};
\node[stage, below=of step3] (step4) {\textbf{Step 4: Submit via Web Interface \faCloudUpload}\\Upload CSVs (batch or individual)\\Provide model name and description};
\node[stage, below=of step4] (step5) {\textbf{Step 5: Automated Evaluation \faCheckCircle}\\Validate submission format → system matches predictions with ground truth\\Compute metrics per dataset};
\node[stage, below=of step5] (step6) {\textbf{Step 6: View Results on Leaderboard \faTrophy}\\Aggregate metrics | Compare across models};

\draw[arrow] (step1) -- (step2);
\draw[arrow] (step2) -- (step3);
\draw[arrow] (step3) -- (step4);
\draw[arrow] (step4) -- (step5);
\draw[arrow] (step5) -- (step6);
\end{tikzpicture}
\caption{Overview of the \textbf{S-GRADES} benchmark user workflow.}
\label{fig:workflow}
\end{figure}

We address this limitation by introducing \textbf{\sgrades} (Studying Generalization of Student Response Assessments in Diverse Evaluative Settings), a unified, open-source benchmark \footnote{\href{https://sgrades.eng.unt.edu}{https://sgrades.eng.unt.edu}} providing standardized evaluation infrastructure through a streamlined workflow (Figure~\ref{fig:workflow}). Our contributions are:

\noindent \textbf{Dataset Aggregation and Standardization.} The benchmark combines $\mathbf{14}$ datasets in AES and ASAG tasks, covering general comprehension, domain-specific assessments, and language proficiency across subjects like English, chemistry, physics, life sciences, and computer science. It includes various response types (persuasive, narrative, expository, source-dependent) from public and proprietary sources. Each dataset undergoes consistent preprocessing and stratified data splits while maintaining original scoring scales to preserve task-specific characteristics (\S \ref{sec:benchmark-datasets}).

\noindent \textbf{Unified Web-Based Evaluation Infrastructure.} \sgrades\ offers a web-based platform \footnote{\href{https://github.com/nlpatunt/sgrades}{\githubicon\ https://github.com/nlpatunt/sgrades}} for dataset access and performance benchmarking. Researchers can download datasets, train models locally, submit predictions, and receive evaluations. The system validates the submissions and calculates multiple metrics (\S \ref{sec:benchmark-devel}).

\noindent \textbf{Systematic Evaluation of State-of-the-Art LLMs.} We evaluate \footnote{\href{https://github.com/nlpatunt/sgrades-experiments}{\githubicon\ https://github.com/nlpatunt/sgrades-experiments}} three frontier LLMs: GPT-4o mini, Gemini 2.5 Flash, and Llama 4 Scout, across six reasoning configurations \cite{liu24colm} (inductive (few-shot) with random sampling, deductive (zero-shot), abductive (Chain-of-Thought), and a combination of them), answering the following questions (\S \ref{sec:experimental-setup}): 

\begin{itemize}[leftmargin=*, itemsep=0pt]
\item \textbf{Model-Reasoning Alignment:} Which LLMs excel with which specific reasoning strategies and why certain approaches align better with their underlying architectures.

\item \textbf{Cross-Dataset and Cross-Paradigm Generalization:} How do model performances change when few-shot exemplars are selected on one dataset and evaluated on another, both within the same paradigm (e.g., \textit{ASAP2.0} (AES) $\rightarrow$ \textit{ASAP-AES} (AES)) and across paradigms (e.g., \textit{ASAP2.0} (AES) $\rightarrow$ \textit{ASAP-SAS} (ASAG)). This tests whether reasoning strategies generalize across fundamentally different response types.

\item \textbf{Exemplar Selection Stability:} How stable are the model predictions across multiple random few-shot sample selections in inductive reasoning setups.
\end{itemize}

Together, these contributions establish a foundation for standardized, reproducible research in LLM-based automated scoring and open new directions for reasoning-aware evaluation.

\section{Related Work}
\label{sec:related_work}
\noindent \textbf{AES. }
Previous work in AES has developed under three paradigms: feature-based statistical models, deep learning, and large language models (LLMs). Early systems, eg.,  PEG \citep{Page1966PEG} and e-rater \citep{Attali2006erater} relied on engineered features capturing surface traits (e.g., word frequency, syntax, sentence length), while IEA introduced latent semantic analysis for semantic similarity \citep{Landauer1998LSA, Foltz1999IEA}. 
Early DNN-based approaches, such as \citet{Taghipour2016neural}, showed that recurrent networks captured writing patterns, and BERT-based models improved contextual understanding \citep{Ludwig2021Transformer}. More recently, \citet{Mizumoto2023potential} and \citet{Xiao2024LLMs} found GPT-4 and Claude closely align with human raters, though sensitive to rubric phrasing. \citet{canAIgrade2024} reported strong correlations ($r=0.87$) for GPT-4, Gemini, and Claude but noted systematic overprediction for mid-range essays. However, a longitudinal evaluation by \citet{Pack2024reliability} showed that GPT-4's reliability dropped from $0.843$ to $0.779$ over $90$ days, revealing instability and bias. These findings highlight persistent challenges in interpretability, reproducibility, and generalizability of LLM-based AES systems.

\noindent \textbf{ASAG. }
\citet{burrows2015eras} identified five methodological eras in ASAG: concept mapping, information extraction, corpus-based similarity, machine learning, and the current evaluation era focused on benchmarks. Early systems like C-rater \citep{burrows2015eras} relied on handcrafted or lexical features.
\citet{haller2022survey} traced the shift from word2vec and GloVe to transformers, noting improved semantic understanding but persistent dataset fragmentation. Domain-specific pretraining of BERT improved grading accuracy by $\approx 10\%$ on benchmarks such as ASAP-SAS \citep{sung2019pre}, while in-context fine-tuning enhanced alignment with human raters \citep{fernandez2022automated}. Despite advances in LLMs, systematic evaluations reveal reliability issues. \citet{henkel2023can} found that rubric-based prompting improved GPT-4 agreement, yet \citet{chang2024automatic} and \citet{grevisse2024llm} reported inconsistent score inflation or deflation across languages and domains. \citet{Schneider2023Autograding} showed ChatGPT's scores were highly prompt-sensitive, and \citet{gao2023grading} found strong binary accuracy but weak partial-credit handling. Studies such as \citet{jiang2024short} (GPT-4), and \citet{xie2024grade} (rubric–review pipeline) show that LLMs can achieve near-human agreement when properly prompted but remain sensitive to rubric design and task length.

Despite recent progress, AES and ASAG still suffer from poor reproducibility and generalization. AES is prone to prompt-specific overfitting and low interpretability, while ASAG is sensitive to domain shifts and rubric changes. Inconsistent dataset sizes and rubrics hinder standardized evaluation, which \sgrades\ helps address. 

\section{\sgrades\ Benchmark Datasets}
\label{sec:benchmark-datasets}
We summarize prior work on the fourteen datasets (Tab. \ref{tab:datasets}) integrated into the platform.

\noindent \textbf{Essay Scoring Studies.}
The \textbf{ASAP-AES} dataset from the 2012 Kaggle competition is a key benchmark for automated essay scoring, foundational for transformer-based AES. Its lack of public test labels limits reproducibility and comparison, as studies rely on cross-validation. \citet{rodriguez2019language} and \citet{ormerod2021automated} set strong transformer baselines, showing both large models (BERT, XLNet) and efficient variants (ALBERT, MobileBERT, Electra) perform reliably with fewer parameters, despite cross-validation's limitations. \citet{mathias2018asap++} expanded ASAP-AES with trait-level annotations to form \textbf{ASAP++}, showing multi-task BiLSTM models enhance holistic score predictions.

\citet{crossley2024large} introduced \textbf{PERSUADE 2.0} with discourse and argumentation annotations on essays, emphasizing rhetorical assessment. Later, \citet{crossley2025ASAP2} introduced \textbf{ASAP 2.0}, a competition dataset with publicly available test labels. Top LLM systems (GPT-4, Claude) achieved 84\% accuracy, though varying metrics (accuracy vs.\ QWK) complicated comparisons with previous AES studies.

\noindent \textbf{Short-Answer Grading Studies.}
\citet{jiang2024short} evaluated prompting strategies on \textbf{ASAP-SAS}, a science short-answer dataset graded by detailed rubrics. They found that exemplar prompts consistently outperformed intermediate reasoning steps (\(\Delta\)QWK = +0.08), though performance varied substantially across question types (QWK range: 0.45--0.82), highlighting task-dependent prompting effectiveness.

\citet{ivanova2024evaluating} tested ChatGPT-3.5 on the \textbf{Mohlar} dataset, which contains introductory computer science answers (e.g., pointers, data structures), graded on a 5-point scale. LLM-human agreement lagged behind human-human levels, especially for technical items, revealing inconsistency in grading CS concepts.

\citet{kortemeyer2024performance} benchmarked GPT-4 on \textbf{SciEntSBank} (multi-domain science questions) and \textbf{BEEtlE} (introductory electricity and circuits), each with 2-way and 3-way classifications. GPT-4 performed strongly on SciEntSBank but struggled with BEEtlE's contradictory cases. Interestingly, omitting reference answers slightly improved accuracy, suggesting reliance on pre-trained knowledge over rubric cues.

\noindent \textbf{Domain-Specific Assessment Studies.}

\citet{sonkar2024automated} introduced the \textbf{Rice\_Chem} dataset for chemistry problem-solving, demonstrating that rubric-aware models with explicit scoring criteria surpass general BERT baselines, though long-form explanations still yield lower human–AI agreement. \citet{xie2024grade} presented the \textbf{OS\_Dataset}, a computer science dataset where students explain OS behavior using code tracing and system reasoning (e.g., registers, threads, memory). This dataset employs a rubric$\rightarrow$grade$\rightarrow$review pipeline that improved grading agreement (87\% vs.\ 76\%). \citet{Feng2024CSEE} studied human–AI scoring with the \textbf{CSEE} dataset, which contains English writing tasks from Chinese high school students, including formal letters, short essays, and argumentative compositions (score range 0--16). This study found that LLM support reduced grading time by 38\% while maintaining reliability, though institutional restrictions limit reproducibility. \citet{gao2024towards} proposed the \textbf{ReGrading dataset (2JC)} for engineering short answers, combining binary and multi-score tasks, showing rubric-based models were accurate but overly lenient.

\noindent \textbf{Language Proficiency Assessments.}
\citet{qiu2024large} studied IELTS writing assessment, showing fine-tuned transformers perform well on argumentative essays, with varying results by prompt type. Comparing GPT-4o and Llama-3, they found that GPT-4o remained stable regardless of exemplars, whereas Llama-3 improved without them, highlighting model-specific optimization needs.

\begin{table*}[!htbp]
\centering
\footnotesize
\renewcommand{\arraystretch}{1.1}
\setlength{\tabcolsep}{3.5pt}
\begin{tabularx}{\textwidth}{l l X l c c}
\toprule
\textbf{Task Type} & \textbf{Dataset} & \textbf{Domain(s)} & \textbf{Score Range(s)} & \textbf{Public} & \textbf{Test Split Size} \\ 
\midrule
\multirow{8}{*}{\textbf{AES}} 
& ASAP-AES & Persuasive, Narrative, and Source-based Essays &
\makecell[l]{Set 1: 2--12;\, Set 2: 1--6;\\ Sets 3--4: 0--3;\\ Sets 5--6: 0--4;\\ Set 7: 0--30;\, Set 8: 0--60} & Yes & 1,298 \\ \cmidrule(l){2-6}
& ASAP++ & Argumentative and Expository Essays & 0--6 & Yes & 1,069 \\ \cmidrule(l){2-6}
& ASAP2.0 & Extended Essay Scoring (Multi-domain) & 0--6 & Yes & 4,946 \\ \cmidrule(l){2-6}
& Persuade\_2 & Persuasive Writing (Argumentative) & 1--6 & Yes & 2,600 \\ \cmidrule(l){2-6}

& IELTS Writing (General) & General English Writing Tasks & 1--9 & Yes & 144 \\ \cmidrule(l){2-6}
& IELTS Writing (Task 2) & Academic English Writing Tasks & 1--9 & Yes & 491 \\ 
\midrule
\multirow{6}{*}{\textbf{ASAG}} 
& ASAP-SAS & Science Short Answers & 0--3 & Yes & 3,409 \\ \cmidrule(l){2-6}
& ReGrading Dataset (2JC) & Conceptual Short-Answer (Engineering, Rubric-Based) & 0--8 & No & 198 \\ \cmidrule(l){2-6}
& CSEE & Short English Writing Tasks (extended answers, letters) & 0--16 & Yes & 2,654 \\ \cmidrule(l){2-6}
& Mohlar & Short Answer Questions (Computer Science concepts) & 0--5 & No & 455 \\ \cmidrule(l){2-6}
& BEEtlE & Physics and Electronics &
\makecell[l]{2-way: 0--1 (Correct/Incorrect);\\ 3-way: 0--2 (Contradictory/\\Correct/Incorrect)} & Yes & 1,258 \\ \cmidrule(l){2-6}
& SciEntSBank & Life Science Conceptual Questions &
\makecell[l]{2-way: 0--1 (Correct/Incorrect);\\ 3-way: 0--2 (Contradictory/\\Correct/Incorrect)} & Yes & 5,835 \\ \cmidrule(l){2-6}
& Rice\_Chem & Chemistry Problem-solving &
\makecell[l]{Q1: 0--8;\, Q2: 0--8;\\ Q3: 0--9;\, Q4: 0--8} & Yes & \makecell[c]{Q1: 66;\, Q2: 66;\\ Q3: 62;\, Q4: 60} \\ \cmidrule(l){2-6}
& OS\_Dataset & Computer Science Conceptual Responses &
\makecell[l]{Q1: 0--19;\, Q2: 0--16;\\ Q3: 0--15;\, Q4: 0--16;\\ Q5: 0--27} & No & \makecell[c]{Q1: 3;\, Q2: 8;\\ Q3: 8;\, Q4: 8;\\ Q5: 8} \\ 
\bottomrule
\end{tabularx}
\caption{Summary of datasets included in the S-GRADES benchmark. \textbf{For datasets that are not currently publicly hosted online, we obtained creator permission for research use and sharing.}}
\label{tab:datasets}
\end{table*}

\section{Benchmark Development}
\label{sec:benchmark-devel}
\subsection{Preprocessing \& Standardization}
We processed datasets into a standard tabular format to ensure unified evaluation while preserving scoring semantics.

\noindent \textbf{Unified Schema. }Each dataset includes three \textbf{core} columns: 
\begin{enumerate}[leftmargin=*, itemsep=0pt]
    \item \textbf{Unique Identifier:} A dataset-specific column such as \texttt{essay\_id}, \texttt{ID}, \texttt{sis\_id}, or \texttt{index} that uniquely identifies each student response.
    \item \textbf{Response Text:} The column containing the student's written response, labeled as \texttt{essay}, \texttt{student\_answer}, or \texttt{response}, depending on the dataset.
    \item \textbf{Ground Truth Score:} The target score column with dataset-specific naming conventions such as \texttt{domain1\_score}, \texttt{label}, \texttt{Score}, or \texttt{band\_score}.
\end{enumerate}
We kept and expanded each dataset with all available context from sources. For datasets with prompt texts, rubric descriptions, essay\_set identifiers, reference materials, or extra metadata, we retained these as supplementary columns for complete context.

\noindent \textbf{Handling Large External Resources.} Some datasets included auxiliary materials such as source essays and images essential for evaluation. These were stored in a separate repository and linked via stable URLs or identifiers, ensuring dataset completeness and efficient distribution.

\noindent \textbf{Preserving Original Score Scales.} We retained each dataset's original scoring scales to respect contextual granularity. Ranges include binary labels (BEEtlE\_2way, SciEntSBank\_2way), extended numeric scales (ASAP-AES Set 8: 0--60, OS Q5: 0--27), and categorical, integer (0--3, 0--8), or band-based (IELTS 1--9) formats. All score definitions are documented.

\noindent \textbf{Data Splits.} We provide standardized data partitions (train/validation/test) for each dataset. Training and validation splits have complete ground-truth labels for development and tuning. Test data is available in two forms: a \textit{public version} (with responses, IDs, and metadata) and a \textit{private version} (with securely stored ground truth labels for evaluation).

\subsection{Platform Architecture}
\sgrades\ provides a web-based evaluation platform built with \verb|FastAPI|. It includes four components: dataset distribution, submission validation, evaluation engine, and leaderboard system.

\noindent \textbf{Dataset Distribution.} All curated datasets are available via a unified download interface, packaging all fourteen datasets with their train/validation/test splits into one archive. Test files keep metadata and identifiers but have ground-truth scores removed. 

\begin{figure}[!htbp]
    \centering
    \includegraphics[width=0.9\linewidth]{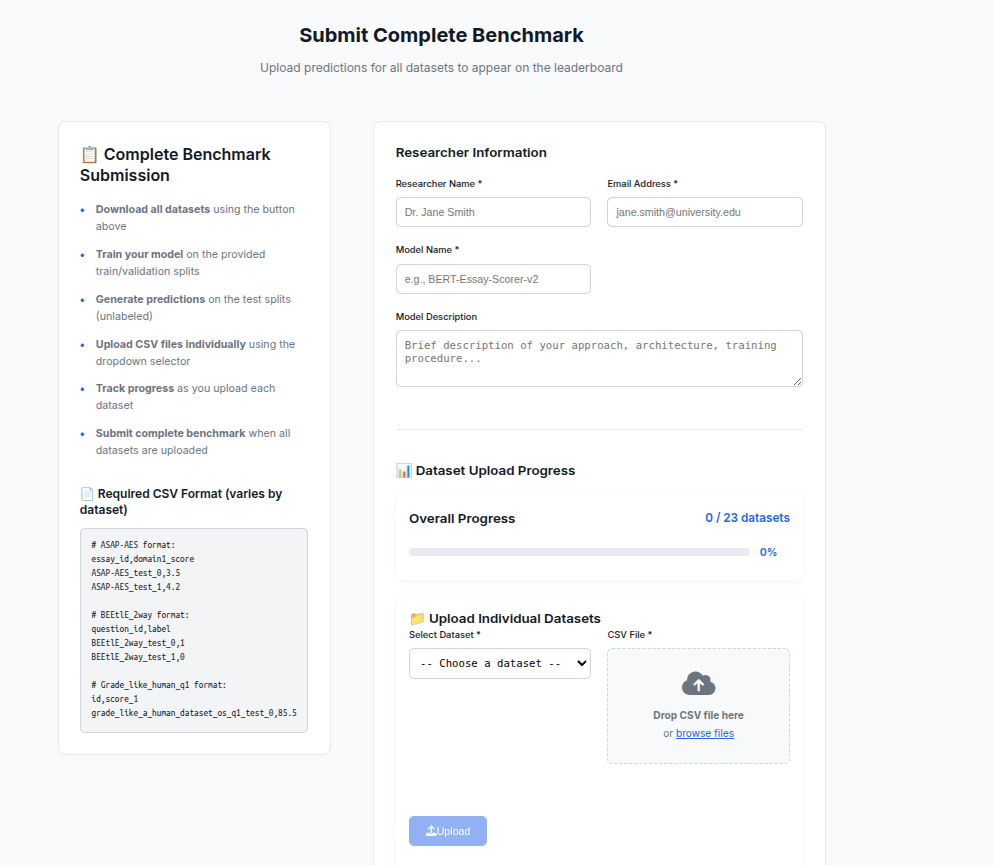}
    \caption{Complete benchmark submission interface.}
    \label{fig:benchmark-submission}
\end{figure}

\noindent \textbf{Submission Validation.} The system includes instructions for uploading individual (per-dataset) and batch (all-dataset) results (Figure \ref{fig:benchmark-submission}). Dataset-specific validator classes conduct multi-stage validation: 1) CSV format verification and security checks to prevent malicious uploads, 2) structural checks for correct column names, 3) ID uniqueness to avoid duplicates, and 4) score format validation confirming predictions match expected types (numeric vs. categorical) and ranges. Errors provide detailed messages on issues and expected formats. 

\noindent \textbf{Evaluation Engine.} Predictions are automatically matched to ground truth by ID, with eight metrics computed for each pair. All results are logged with submission metadata (timestamp, IP, file hash) to ensure reproducibility and integrity.

\begin{figure}[!htbp]
    \centering
    \includegraphics[width=0.9\linewidth]{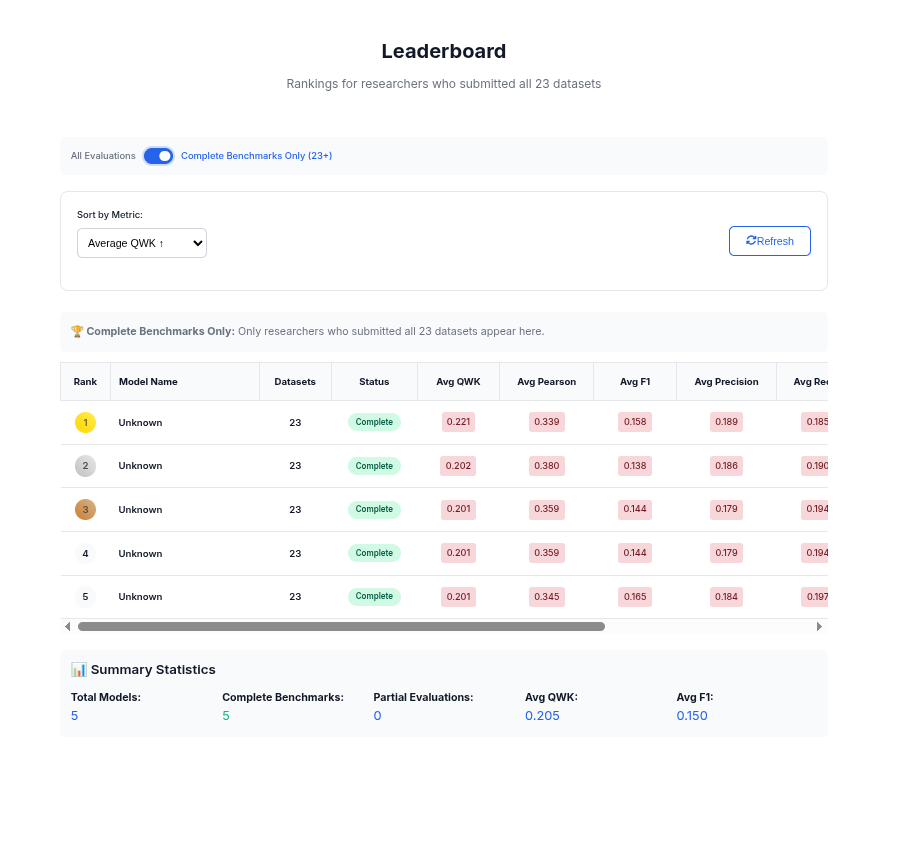}
    \caption{Public leaderboard displaying aggregated results across all datasets and evaluation metrics.}
    \label{fig:leaderboard}
\end{figure}

\noindent \textbf{Leaderboard System.} Results are aggregated and shown on a public leaderboard (Figure \ref{fig:leaderboard}), supporting filtering by metrics like QWK, Pearson correlation, and F1. It displays overall performance across datasets, allowing researchers to compare rankings. 

\noindent \textbf{Security Measures.} Platform security is implemented with rate limits on submissions (5 per minute for single uploads and 2 for batch), and SQL injection attacks are prevented using SQLAlchemy ORM's parameterized queries.

\subsection{Evaluation Metrics}
\sgrades\ supports the computation of the following metrics:

\noindent \textbf{Agreement Metrics.} Quadratic Weighted Kappa (QWK) \cite{McHugh2012} serves as the primary agreement measure, capturing inter-rater reliability between model predictions and human scores while accounting for the ordinal nature of grading scales. It penalizes larger discrepancies more heavily, rewarding models that maintain near-miss consistency.
{\small
\begin{equation*}
    \text{QWK} = 1 - \frac{\sum_{i,j} w_{ij} O_{ij}}{\sum_{i,j} w_{ij} E_{ij}},
\end{equation*}
}
where $O_{ij}$ and $E_{ij}$ are the observed and expected rating matrices, and $w_{ij} = \frac{(i-j)^2}{(k-1)^2}$ penalizes larger score differences across $k$ possible score levels. In contrast, the Pearson correlation coefficient (\textit{r}) measures the strength of the linear relationship between predicted ($\hat{y}_i$) and true ($y_i$) scores.
{\small
\begin{equation*}
    r = \frac{\sum_{i=1}^{N} (y_i - \bar{y})(\hat{y}_i - \bar{\hat{y}})}
    {\sqrt{\sum_{i=1}^{N} (y_i - \bar{y})^2} \sqrt{\sum_{i=1}^{N} (\hat{y}_i - \bar{\hat{y}})^2}},
\end{equation*}
}

\noindent \textbf{Error Metrics.} MAE measures average prediction error in the original scoring scale, while RMSE highlights significant errors, identifying models with consistent over/underestimation.
{\small
\begin{equation*}
\text{MAE} = \frac{1}{N} \sum_{i=1}^{N} |y_i - \hat{y}_i| \quad
\text{RMSE} = \sqrt{\frac{1}{N} \sum_{i=1}^{N} (y_i - \hat{y}_i)^2}
\end{equation*}
}

\noindent \textbf{Classification-Oriented Metrics.} For datasets modeled as classification tasks (\textbf{BEEtlE}, \textbf{SciEntSBank}), we report accuracy, F1, precision, and recall. For regression datasets, predictions and true scores are rounded to the nearest integer class before computation.

These metrics provide a balanced evaluation of automated scoring performance. QWK captures ordinal agreement, Pearson's $r$ reflects rank consistency, and MAE/RMSE quantify absolute and squared errors on original scales. F1, precision, recall, and accuracy assess categorical correctness and, together, reveal both ranking reliability and calibration quality.

\section{Experimental Setup}
\label{sec:experimental-setup}

\noindent \textbf{Models. }We conduct a comprehensive evaluation across three state-of-the-art large language models: Llama-4-Scout (405B, open weight), Gemini 2.5 Flash (proprietary), and GPT-4o-mini (est. $<$30B, proprietary), representing different architectures and scale. All experiments were run through OpenRouter \footnote{\href{https://openrouter.ai/}{https://openrouter.ai/}} with the decoding temperature set to $0.1$ for the main evaluation runs. The experiments yielded $\approx \mathbf{37K}$ scored responses from 14 datasets, covering short factual and long-form argumentative tasks. 

\noindent \textbf{Reasoning Configurations. }We evaluate six reasoning configurations (as categorized in \citet{liu24colm}) covering individual and hybrid approaches.
The \textit{Deductive Reasoning (principle-based application)} is a zero-shot setup in which the model is provided with general scientific or scoring principles. Prompts include a few examples illustrating how principles lead to specific outcomes (e.g., correct, partial, or incorrect), but no labeled examples are provided. The \textit{Inductive Reasoning (5 exemplars)} is a few-shot setup where the model learns grading patterns from five labeled examples \textit{sampled from the training split of the same dataset}. The \textit{Abductive reasoning (inference to the best explanation)} is similar to a Chain-of-Thought prompting setup, in which the models are asked to generate hypotheses. LLMs suggest grading explanations (partial understanding, misconception, correct reasoning), identify the most plausible one, and determine the correct classification or score. \textit{Hybrid} modes combine two reasoning configurations. For example, \textit{Ind + Ded} $\implies$ the prompt includes both scoring principles and exemplars (see Appendix~\ref{appendix:prompts} for detailed descriptions and examples).

\noindent \textbf{Prompt template.} Each reasoning strategy follows a structured, multi-part template ensuring consistency across models and datasets. It begins with a \textbf{system role} defining the evaluator's perspective (e.g., ``You are an expert evaluator using \textit{inductive reasoning}''), followed by detailed \textbf{reasoning instructions} outlining the step-by-step process (such as learning from examples, applying principles, or inferring explanations). \textbf{Illustrative examples} are then provided: five exemplars for inductive prompts; or 2–3 reasoning demonstrations for deductive and abductive prompts based on scientific or scoring principles. The \textbf{evaluation task} presents the target question and student answer to be classified or scored, and an explicit \textbf{output constraint} directs the model to produce only the final label or numeric score without justification (see Appendix~\ref{appendix:prompts} for full prompt templates).\\
\noindent \textbf{Ablation Experiments.} 
Two sources of variability exist in this experimental setup: (1) \textit{exemplar selection}, where inductive setups randomly sample training examples that directly influence the prompt, and (2) \textit{decoding randomness}, introduced by non-zero temperature. To isolate each source, we design two separate stability experiments. In the \textbf{Prediction Stability} experiment, we query each model three times on a subset of the test data, using a fixed set of exemplars in the inductive setups and a fixed temperature of 0.1 for all reasoning setups.
The \textbf{Exemplar Selection Stability} experiments assess the impact of random exemplar selection in inductive setups by keeping the temperature fixed at $0$, but selecting different examples for each run. Finally, we test generalization in \textbf{In/Cross-Paradigm Generalization} experiment where we select representative datasets with consistent performance from each paradigm (ASAP2.0 and ASAP-AES for AES, ASAP-SAS and CSEE for ASAG), and evaluated the effect of using exemplars from one dataset for another dataset (in the inductive setups).

\section{Results and Discussion}
\label{sec:results}
ASAG tasks differ in their output formulation; we distinguish between \textbf{regression-based} and \textbf{classification-based} setups. While we compute all appropriate evaluation metrics, we report QWK as the primary metric, as it is most widely adopted.

\begin{figure}[!htbp]
    \centering
    \includegraphics[width=\linewidth, scale=0.5]{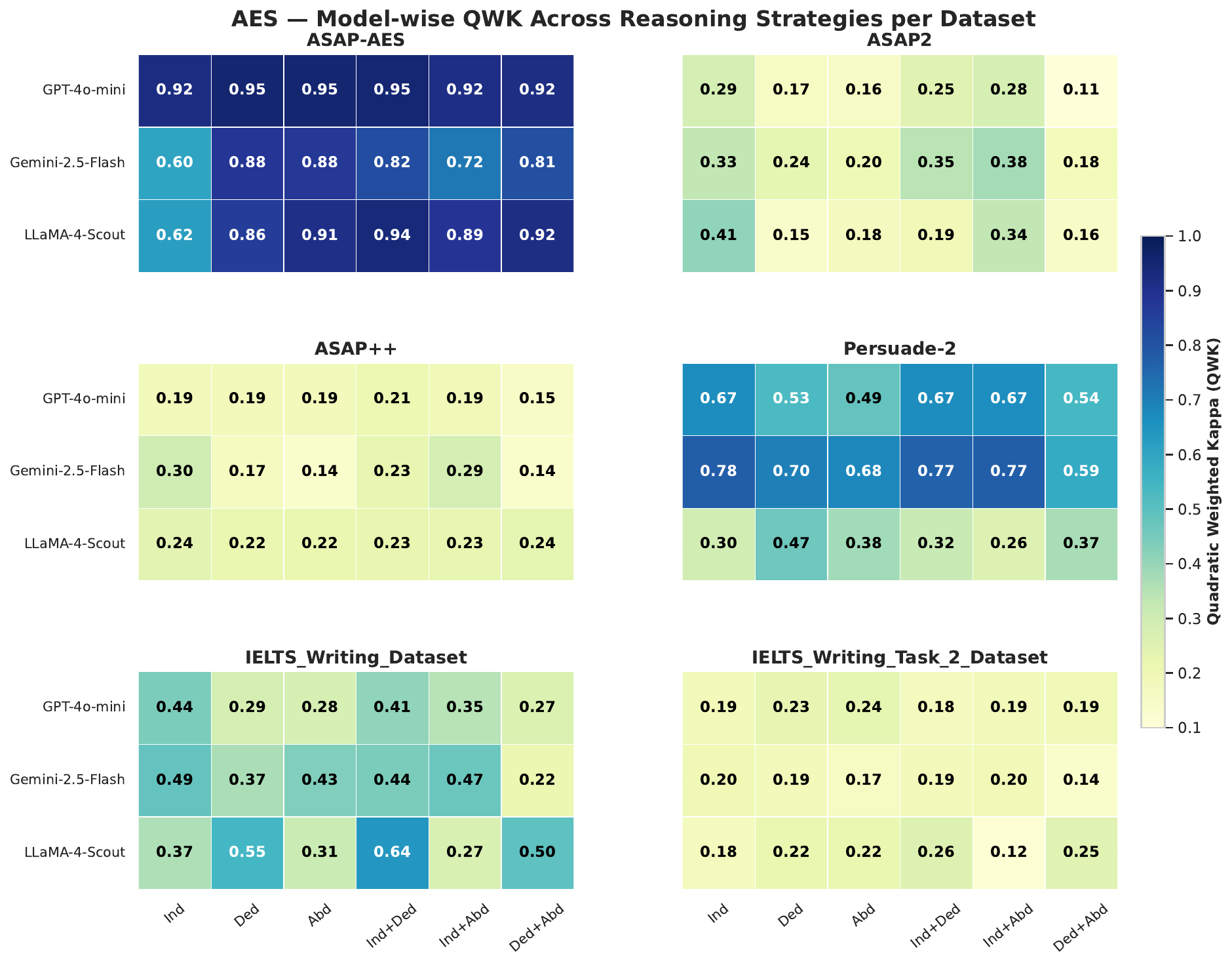}
    \caption{
    QWK scores for AES datasets across models and reasoning strategies.}
    \label{fig:aes_modelwise_heatmaps}
\end{figure}

\noindent \textbf{AES Tasks.} Figure \ref{fig:aes_modelwise_heatmaps} summarizes the QWK performance. Scores range from 0.10 (\textit{GPT-4o-mini}, \textit{ASAP++}, \textit{Ded+Abd}) to 0.96 (\textit{GPT-4o-mini}, \textit{ASAP-AES}), showing wide variation driven by dataset complexity, reasoning type, and model capability.

Across models and datasets, clear performance patterns emerge. GPT-4o-mini shows the highest internal consistency, reaching near-perfect agreement on ASAP-AES but declining on ASAP++ and ASAP2.0, indicating sensitivity to dataset structure. Gemini-2.5-Flash performs most evenly, excelling on Persuade-2, while LLaMA-4-Scout exhibits selective strengths but struggles on most datasets except ASAP-AES.

At the dataset level, ASAP-AES is the most predictable, with all models achieving QWK above 0.75, while ASAP++ is the most difficult due to prompt and rubric diversity. ASAP2.0 shows moderate complexity, Persuade-2 displays large model-dependent variation, and the IELTS datasets show medium difficulty reflecting stricter evaluation criteria and nuanced linguistic expectations typical of standardized language proficiency tasks.

Reasoning strategy proves the most influential factor. Hybrid modes outperform single strategies: Ind+Ded consistently leads, Ind+Abd excels in persuasive writing tasks, and Ded+Abd performs the worst. Among single modes, inductive reasoning provides the most stable baseline. GPT-4o-mini is the most stable across reasoning strategies, Gemini-2.5-Flash maintains moderate but consistent transferability, and LLaMA-4-Scout shows high variance.

\begin{figure}[!htbp]
    \centering
    \includegraphics[width=\linewidth]{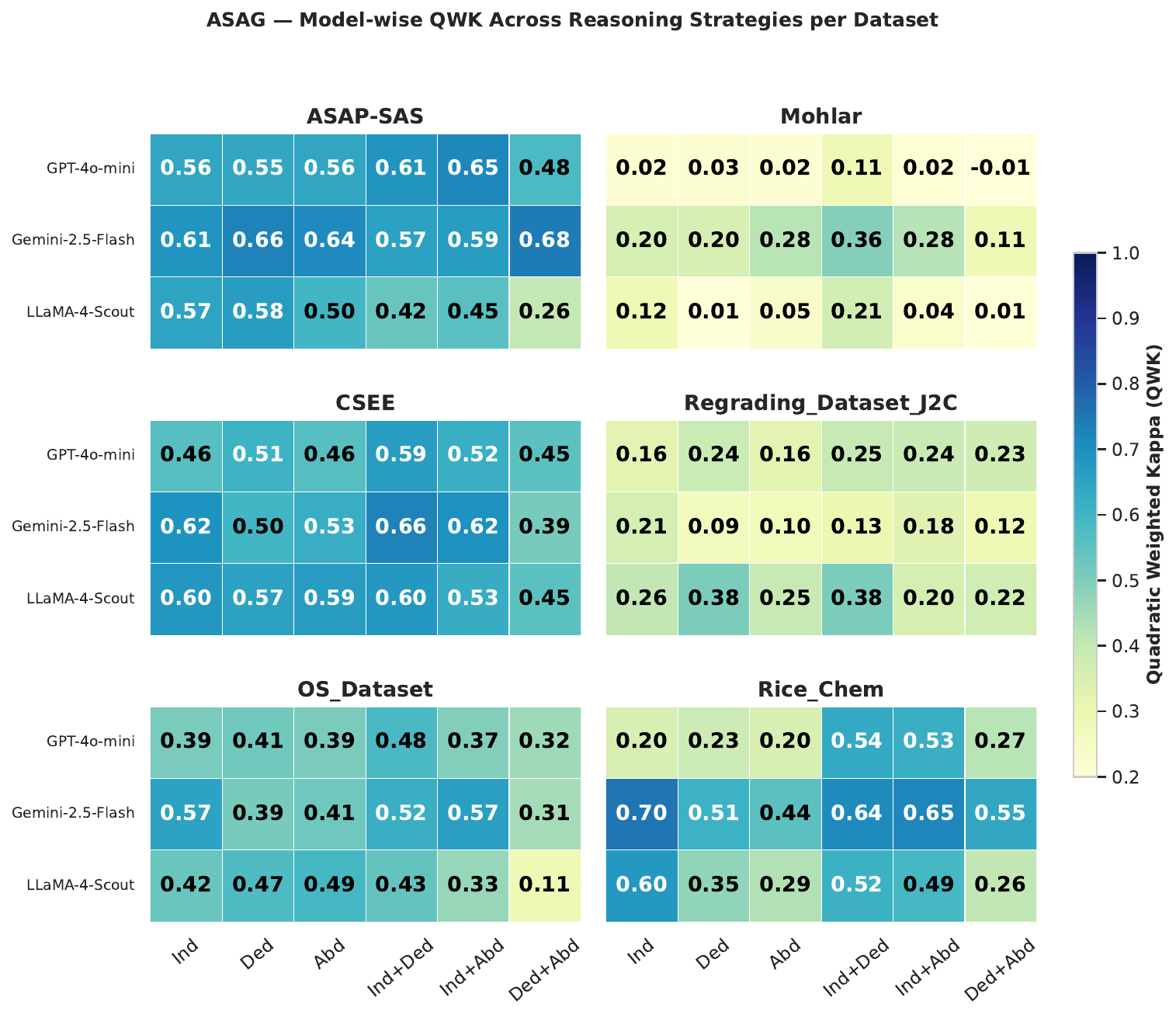}
    \caption{QWK scores for ASAG regression datasets across models and reasoning strategies.}
    \label{fig:asag_heatmaps}
\end{figure}

\noindent \textbf{ASAG Regression.} Figure \ref{fig:asag_heatmaps} presents QWK results for six ASAG regression datasets. Compared to AES tasks, ASAG results show greater variability and lower absolute performance, reflecting the greater difficulty of short-answer grading.

As in AES, GPT-4o-mini performs reliably on structured datasets such as ASAP-SAS and CSEE, indicating strong alignment with explicit scoring rubrics. Gemini-2.5-Flash demonstrates the most balanced performance, leading on Rice\_Chem and maintaining competitive results across domains, suggesting robust generalization. In contrast, LLaMA-4-Scout shows moderate capability on structured tasks but weaker results on short-context datasets like Mohlar, reflecting sensitivity to input length and ambiguity.

At the dataset level, ASAP-SAS remains the most stable benchmark, while Mohlar is the most difficult due to short, under-specified responses. Rice\_Chem achieves mid-to-high performance, consistent with its structured, scientific nature. In contrast, CSEE, although classified as an ASAG dataset, contains short-form writing tasks that resemble essay responses. This essay-like structure likely contributes to its relatively stronger performance compared to other ASAG datasets. ReGrading Dataset (2JC) shows the lowest cross-model stability, highlighting the complexity of revision-based scoring (see Appendix~\ref{appendix:strategy_sensitivity} for detailed dataset characteristics 
and statistics).

Reasoning strategy remains the primary determinant of performance. The hybrid Ind+Ded approach performs best overall, while Inductive reasoning remains a strong baseline for factual and definition-based tasks. The Ded+Abd strategy consistently underperforms, confirming its unsuitability for short-answer grading. In terms of robustness, Gemini-2.5-Flash exhibits the lowest cross-strategy variance ($\sigma \!<\! 0.06$), GPT-4o-mini shows moderate fluctuations, and LLaMA-4-Scout displays the highest sensitivity.

\noindent \textbf{AES vs.\ ASAG Regression.} We averaged results across all AES and ASAG regression datasets to capture overall task difficulty. While AES focuses on writing quality, ASAG emphasizes factual accuracy and rubric alignment, thus being more sensitive to reasoning quality. AES tasks have higher agreement (mean QWK $\approx$ 0.42–0.43 across models) than ASAG ($\approx$ 0.34–0.43), illustrating the challenges of grading short answers with limited context. \textbf{Gemini-2.5-Flash} performs consistently across both domains (QWK = 0.433 each), \textbf{GPT-4o-mini} excels in essays, and \textbf{LLaMA-4-Scout} is less stable. 

\begin{figure}[!htbp]
    \centering
    \includegraphics[width=0.95\linewidth]{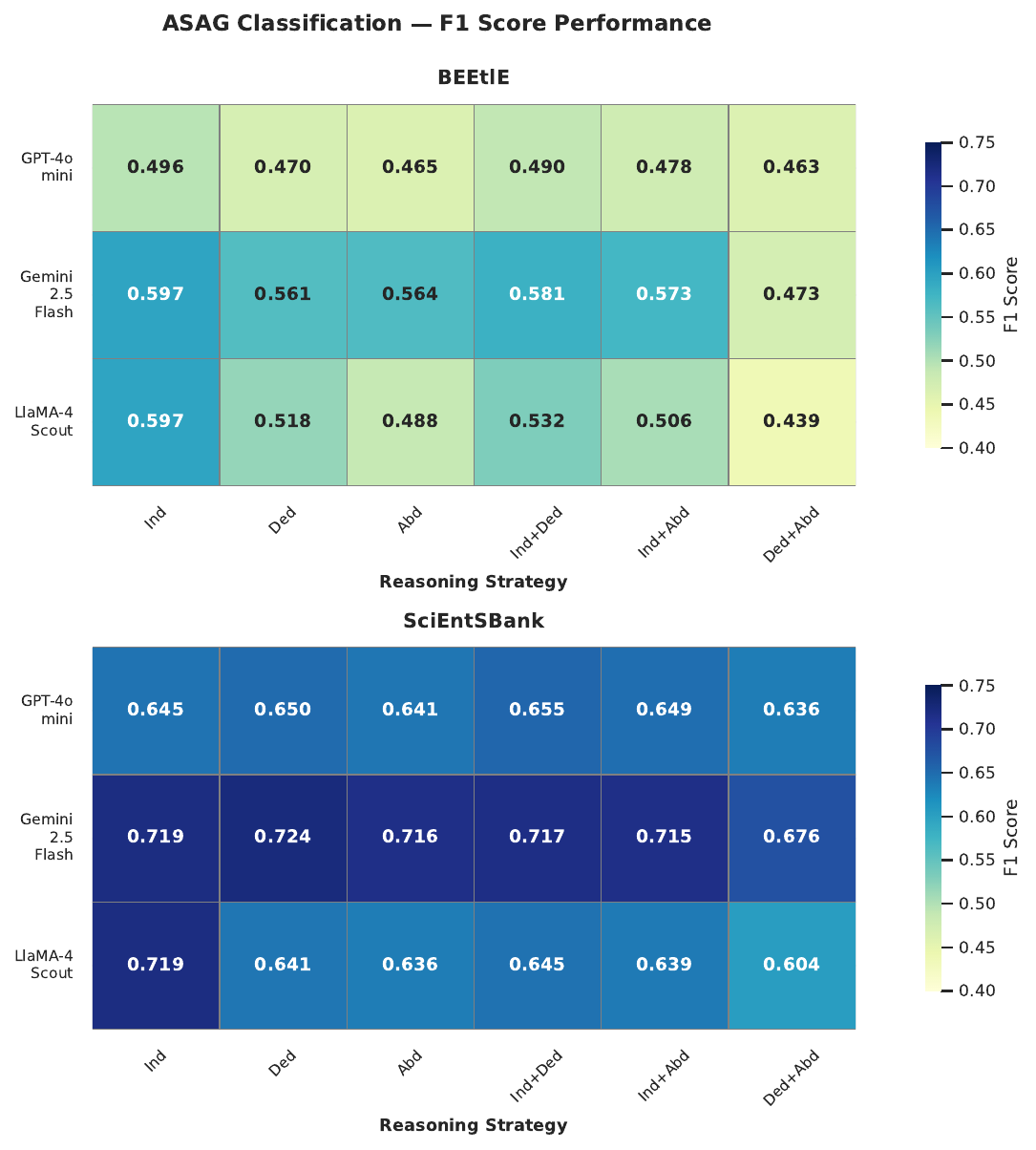}
    \caption{F1 scores across reasoning strategies for 2 classification-based ASAG datasets.}
    \label{fig:asag_modelwise_heatmaps}
\end{figure}

\noindent \textbf{ASAG Classification.} For two ASAG datasets, BEEtlE and SciEntSBank, answers are labeled as correct/incorrect (2-way) or correct/incorrect/contradictory (3-way). As shown in Figure \ref{fig:asag_modelwise_heatmaps}, classification yields more stable metrics than regression, indicating models more reliably distinguish correct from incorrect responses. Across datasets, Gemini-2.5-Flash achieves the highest F1 (0.72), surpassing GPT-4o-mini and LLaMA-4-Scout. Hybrid reasoning strategies (\textit{Ind+Ded}, \textit{Ind+Abd}) provide small but consistent gains.

\subsection{Prediction Stability} To evaluate whether model and strategy rankings are robust to repeated inference, we ran each model three times on a 21\% subset of each dataset's test split.  All experiments were conducted with the temperature set to 0.1, consistent with the original experimental setup.

For AES, the results are highly stable: model ranking and reasoning ranking change only slightly across dataset–strategy combinations. For ASAG regression, model rankings remain stable across datasets, with minor score fluctuations on \textit{OS\_Dataset} and \textit{Regrading\_Dataset\_2JC}, where single-call estimates are inherently noisy due to small test sizes. For ASAG classification, model rankings remain largely consistent across both settings. Gemini-2.5-Flash leads across nearly all conditions, with the sole consistent exception being \textit{Ded+Abd} on BEEtlE, where GPT-4o-mini marginally outperforms it. Overall, these observations indicate that the benchmark's main conclusions remain qualitatively consistent under repeated querying. Detailed per-call variance analysis is reported in Appendix~\ref{app:stability}.

\subsection{Exemplar Selection Stability}\label{sec:exemplar_stability} In inductive setups, exemplars are randomly selected. To test robustness to exemplar variability, we reran the evaluations on ASAP-AES with three random seeds (42, 123, 456) and a fixed temperature of 0, ensuring that all observed variance arises exclusively from exemplar selection rather than from decoding randomness. As shown in Table~\ref{tab:seed_stability}, Gemini-2.5-Flash achieves both the highest mean QWK (0.9520) and the greatest stability (QWK Std = 0.0047), demonstrating that its performance is largely invariant to which examples are shown. GPT-4o-mini (mean QWK = 0.9397, Std = 0.0099) and LLaMA-4-Scout (mean QWK = 0.9450, Std = 0.0111) exhibit slightly higher sensitivity to exemplar choice. Across all three seeds, QWK scores remain consistent within each model, suggesting that exemplar selection does not substantially alter evaluation outcomes under the inductive reasoning setup.

\begin{table}[h]
\centering
\footnotesize
\renewcommand{\arraystretch}{1.1}
\setlength{\tabcolsep}{8pt}
\begin{tabular}{l c}
\toprule
\textbf{Model} & \textbf{QWK Std} \\
\midrule
Gemini-2.5-Flash  & 0.0047 \\
GPT-4o-mini       & 0.0099 \\
LLaMA-4-Scout     & 0.0111 \\
\bottomrule
\end{tabular}
\caption{Exemplar sampling stability on ASAP-AES across three random seeds (42, 123, 456) at temperature = 0. Low standard deviations indicate that evaluation outcomes are largely invariant to the selection of examples.}
\label{tab:seed_stability}
\end{table}

\subsection{Exemplar Generalization} We want to examine whether LLM reasoning abilities transfer in the inductive setups by selecting five samples from one dataset (base) and using them as exemplars in another dataset (target). Depending on whether the base and target datasets come from the same task family (ASAG/AES) or different ones, we refer to it as the ``In'' or ``Cross'' paradigm of generalization.

\begin{table}[!t]
\centering
\small
\setlength{\tabcolsep}{0pt}
\begin{tabular}{lcc}
\toprule
\textbf{Model} & \textbf{Reasoning} & \textbf{$\Delta$QWK\%} \\
\midrule
\textit{In-paradigm generalizations} \\
\textbf{AES:} \textit{ASAP2.0 $\rightarrow$ ASAP-AES} \\
GPT-4o-mini & Ind+Ded & -5.3 \\
LLaMA-4-Scout & Ind+Ded & -96.2 \\
Gemini-2.5-Flash & Ind+Ded & +9.4 \\
\midrule
\textbf{ASAG}: \textit{ASAP-SAS $\rightarrow$ CSEE} \\
GPT-4o-mini & Ind+Abd & -13.4 \\
LLaMA-4-Scout & Ind+Ded & -17.0 \\
Gemini-2.5-Flash & Ind & -9.0 \\
\midrule
\textit{Cross-paradigm generalizations} \\
\textbf{ASAG-AES:} \textit{CSEE $\rightarrow$ ASAP-AES} \\
GPT-4o-mini & Ind+Ded & -2.8 \\
LLaMA-4-Scout & Ind+Ded & -95.5 \\
Gemini-2.5-Flash & Ind+Ded & +12.5 \\
\midrule
\textbf{AES-ASAG:} \textit{ASAP-AES $\rightarrow$ CSEE}\\
\hspace*{3.2em}\textit{ASAP-SAS}\\
GPT-4o-mini & Ind+Abd & -13.2 \\
LLaMA-4-Scout & Ind & -47.2 \\
Gemini-2.5-Flash & Ind & -75.5 \\
\midrule
\end{tabular}
\caption{Exemplar generalization results. For inductive (5-shot) reasoning setups, exemplars are selected from one dataset and used on another.}
\label{tab:exemplar-gen}
\end{table}

Table \ref{tab:exemplar-gen} presents the results (full results across 
all datasets in Appendix~\ref{appendix:generalization}). We use the best-performing dataset from prior evaluations as the base for most cross-generalization experiments and, for each transfer, apply the reasoning strategy that performs best on this base dataset. Within the AES in-paradigm setting, GPT-4o-mini exhibits a minor performance decline, Gemini-2.5-Flash shows an unexpected improvement, and LLaMA-4-Scout experiences a significant collapse, indicating instability in transferring scoring behavior across similar essay-scoring datasets. For the ASAG in-paradigm condition, all models exhibit moderate performance degradation under cross-dataset transfer, while retaining reasonable grading capability (QWK > 0.49). This suggests that, although models' performance is dataset-specific, their underlying reasoning partially generalizes within the short-answer domain. 
Interestingly, in the cross-paradigm transfer from ASAG to AES, exemplars drawn from the CSEE dataset yield measurable performance gains on ASAP-AES for the Gemini-2.5-Flash model, suggesting that certain structured, form-oriented writing exemplars, such as formal letters and compositions, may positively scaffold the evaluation of student responses for this model. In contrast, transfers from AES to ASAG consistently reduce performance across all models, aligning with expectations that rubric-driven essay exemplars offer limited utility for factual, short-answer evaluation tasks. Reasoning configuration affects performance, but is not the dominant factor.

\section{Conclusion and Future Work}
We introduce \sgrades, a unified and extensible benchmark that consolidates 14 AES and ASAG datasets comprising over 37K graded student responses under a standardized evaluation framework. By enabling systematic and reproducible comparison across grading paradigms, \sgrades\ addresses long-standing fragmentation in educational NLP and provides a foundation for studying cross-task generalization.

Using the benchmark as a testbed, we evaluate three state-of-the-art LLMs under multiple reasoning configurations and analyze the impact of exemplar selection and transfer. These experiments illustrate how standardized, cross-dataset evaluation surfaces meaningful differences in model robustness, reasoning sensitivity, and grading consistency. In particular, the results highlight persistent performance gaps between AES and ASAG, underscoring the need for a unified evaluation framework when assessing model generalization in student response grading.

Future work will extend \sgrades\ to multilingual and multimodal settings, improve exemplar selection stability, and incorporate additional reasoning and rubric-grounded prompting strategies. We hope \sgrades\ serves as a community resource for rigorous, transparent, and extensible evaluation of automated student assessment systems.

\section{Limitations}

While \sgrades\ offers the most comprehensive reasoning-based evaluation to date, several limitations remain that motivate future directions. 

\begin{itemize}[leftmargin=*, itemsep=0pt]
    \item \textbf{Model Scope:} Our experiments included only three models: \textit{GPT-4o-mini}, \textit{Gemini-2.5-Flash}, and \textit{LLaMA-4-Scout}. Expanding coverage to newer open-weight and proprietary models (e.g., Llama~3.1, Qwen~2.5, Mistral~Large) will help validate architectural trends and reasoning scalability.
    
    \item \textbf{Exemplar Variance:} Performance varied substantially with random seed selection. Future studies should replicate across multiple random seeds and explore structured sampling to reduce noise and better isolate genuine reasoning gains.
    
    \item \textbf{Task Coverage:} Although S-GRADES consolidates 14 datasets, all are English and text-only. Extending to multilingual, multimodal, and non-Western educational data is critical for assessing generalization beyond current benchmarks.

    \item \textbf{ASAG Domain Specificity.} Short-answer grading remains far harder than essay scoring (mean QWK: 0.34--0.41 vs.\ 0.42--0.43), driven by rubric and question variability. Even GPT-4o-mini's best cross-domain drop (-13.2\%) exceeds AES losses (-2.8\%). Improving ASAG transfer likely requires rubric-aware training, domain-specific pretraining, or hybrid human–AI grading with educator validation.

    \item \textbf{Methodological Extensions:} Future work should explore richer reasoning combinations (e.g., Inductive + Deductive + Abductive) to test whether multi-step reasoning can enhance generalization across datasets. Integrating these with fine-tuning or meta-learning may help mitigate ASAG transfer failures, especially through rubric-aware or normalized scoring methods.

\end{itemize}

Overall, these directions aim to advance reproducibility, reduce sampling artifacts, and enable more generalizable educational evaluation with reasoning-augmented language models.

\section*{References}
\bibliographystyle{latex/lrec2026-natbib}
\bibliography{latex/lrec2026-example}

\newpage
\appendix
\section{Comprehensive Reasoning Strategy Analysis}
\label{appendix:reasoning_analysis}

This appendix provides detailed sensitivity analysis across all reasoning configurations, models, and datasets evaluated in S-GRADES. We quantify the extent to which performance varies with reasoning-strategy selection, complementing the primary results reported in Section~\ref{sec:results}.

\subsection{Strategy Sensitivity Metrics}
\label{appendix:strategy_sensitivity}

To measure sensitivity to reasoning strategy selection, we compute the standard deviation ($\sigma$) of QWK scores across the six reasoning strategies (Ind, Ded, Abd, Ind+Abd, Ind+Ded, Ded+Abd) for each dataset-model pair. A higher $\sigma$ indicates that performance is more sensitive to strategy choice, while a lower $\sigma$ indicates consistent performance regardless of strategy.\\

Formally, for a model $m$ on dataset $d$, the per-dataset strategy variance is:
\begin{equation}
\sigma_{m,d} = \sqrt{\frac{1}{6}\sum_{s=1}^{6}(\text{QWK}_{m,d,s} - \overline{\text{QWK}}_{m,d})^2}
\end{equation}

where $\overline{\text{QWK}}_{m,d}$ is the mean QWK across all six strategies for model $m$ on dataset $d$.\\

To summarize overall strategy sensitivity for a model across all datasets, we compute the mean strategy variance:
\begin{equation}
\sigma_{\text{strategy}} = \frac{1}{D}\sum_{d=1}^{D}\sigma_{m,d}
\end{equation}

where $D$ is the number of datasets in the task type. Lower $\sigma_{\text{strategy}}$ indicates greater stability across reasoning strategies overall.\\

\begin{figure}[!htbp]
\centering
\includegraphics[width=\columnwidth]{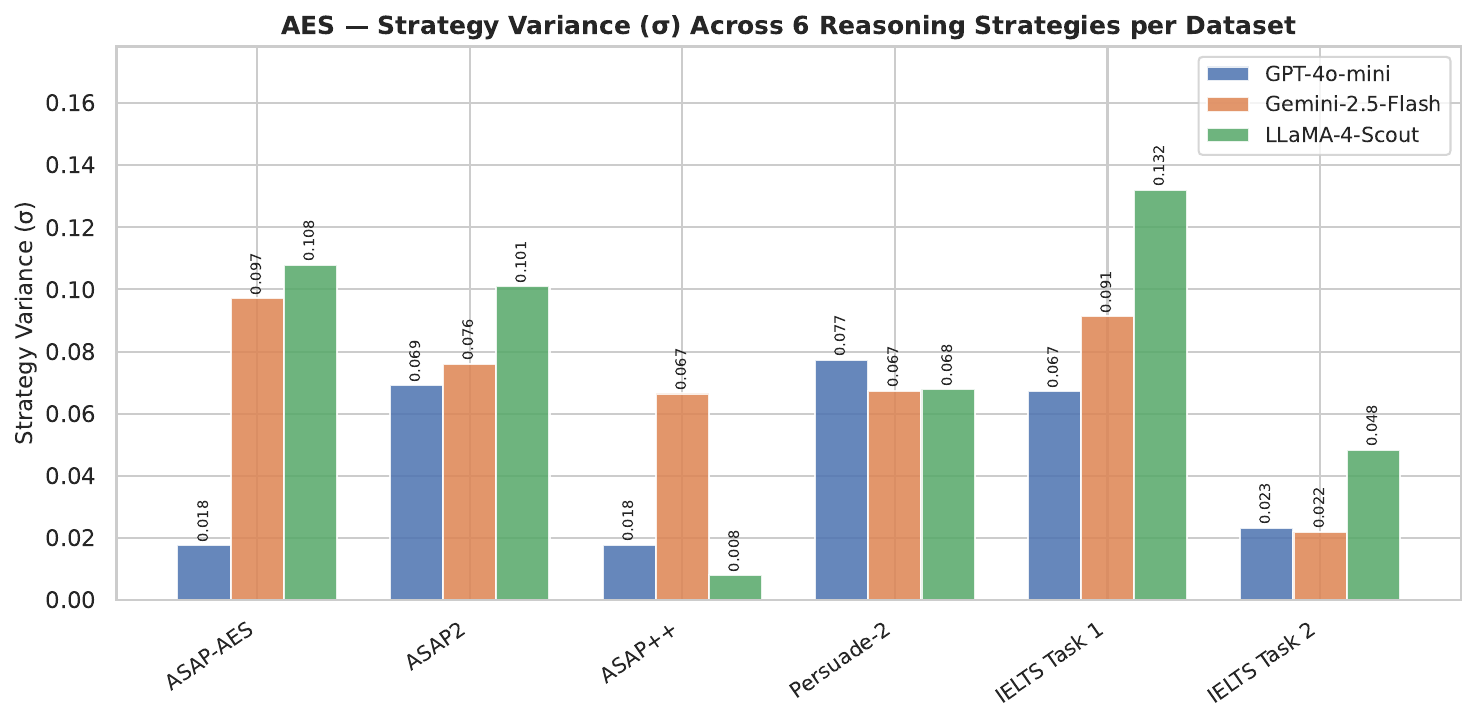}
\caption{Strategy variance ($\sigma_{m,d}$) across 6 reasoning strategies per AES dataset. Higher values indicate greater sensitivity to reasoning strategy selection.}
\label{fig:strategy_variance_aes}
\end{figure}

\begin{figure}[!htbp]
\centering
\includegraphics[width=\columnwidth]{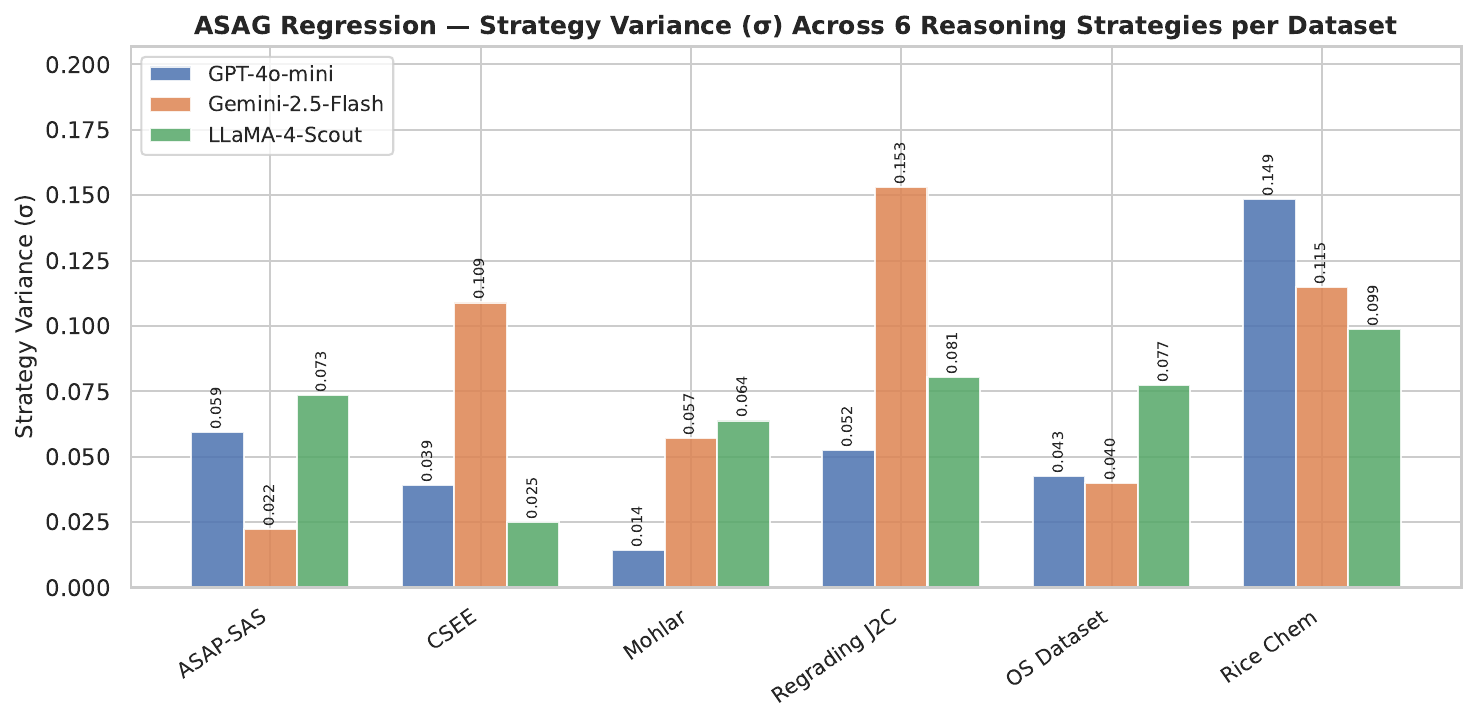}
\caption{Strategy variance ($\sigma_{m,d}$) across 6 reasoning strategies per ASAG regression dataset. Higher values indicate greater sensitivity to the selection of reasoning strategies.}
\label{fig:strategy_variance_asag}
\end{figure}

Figures~\ref{fig:strategy_variance_aes} and~\ref{fig:strategy_variance_asag} 
report per-dataset $\sigma_{m,d}$ values (Equation 1) for AES and ASAG 
regression tasks respectively.

To visually illustrate the per-dataset variance patterns, Figures~\ref{fig:strategy_variance_aes} and~\ref{fig:strategy_variance_asag} plot the per-dataset variance patterns as grouped bar charts. Each group represents one dataset, with bars corresponding to the three models. The figures were generated by computing $\sigma_{m,d}$ directly from the QWK scores in our experimental results using Equation~1, with each bar height representing how much a model's performance fluctuates across the six reasoning strategies on that dataset.

\textbf{Key Observations.}
For AES tasks, GPT-4o-mini exhibits the lowest average strategy variance ($\sigma_{\text{strategy}} = 0.045$), indicating robust performance across reasoning configurations. This stability is most pronounced on ASAP-AES ($\sigma_{m,d} = 0.018$) and ASAP++ ($\sigma_{m,d} = 0.018$), where performance remains consistent across all six strategies. Gemini-2.5-Flash shows moderate sensitivity ($\sigma_{\text{strategy}} = 0.070$), while LLaMA-4-Scout exhibits the highest variance ($\sigma_{\text{strategy}} = 0.078$), particularly on IELTS Task 1 ($\sigma_{m,d} = 0.132$) where strategy selection critically impacts evaluation quality.

For ASAG regression, the pattern reverses: Gemini-2.5-Flash becomes the most strategy-sensitive model ($\sigma_{\text{strategy}} = 0.083$), with extreme variance on ReGrading Dataset (2JC) ($\sigma_{m,d} = 0.153$) driven by the catastrophic failure of pure inductive reasoning (QWK $= -0.20$). GPT-4o-mini maintains the lowest average variance ($\sigma_{\text{strategy}} = 0.060$) for structured datasets such as ASAP-SAS and CSEE, while LLaMA-4-Scout shows moderate sensitivity ($\sigma_{\text{strategy}} = 0.070$). These contrasting patterns suggest that strategy sensitivity is both model-dependent and task-dependent, with no single model consistently robust across all evaluation scenarios.

\textbf{Model Stability Rankings:} Across all tasks, GPT-4o-mini exhibits the smallest strategy variance (AES: $\sigma = 0.045$; ASAG Regression: $\sigma = 0.060$), indicating consistent performance regardless of reasoning approach. Gemini-2.5-Flash 
shows moderate variance (AES: $\sigma = 0.070$; ASAG Regression: $\sigma = 0.083$) with occasional catastrophic failures (ReGrading Dataset (2JC) \textit{Ind}: QWK $= -0.20$). LLaMA-4-Scout exhibits the highest variance in AES ($\sigma = 0.078$), particularly on IELTS Task 1 (range: 0.27--0.64).

\textbf{ASAG Classification Strategy Variance.} 
Figure~\ref{fig:asag_modelwise_heatmaps} presents F1 scores for BEEtlE and SciEntSBank across all reasoning strategies. Unlike AES and ASAG regression tasks, classification datasets show narrower score ranges within each model (GPT: 0.27--0.44; Gemini: 0.21--0.52; LLaMA: 0.14--0.40), indicating moderate sensitivity to reasoning strategy selection. Notably, \textit{Ded+Abd} consistently achieves the lowest F1 across both datasets and all models, while individual \textit{Ind} and \textit{Ded} strategies match or exceed hybrid approaches — a pattern distinct from AES tasks where hybrid strategies provide substantial gains.

\section{Dataset Specific Anomalies}
\label{appendix:anomalies}

The following cases represent extreme strategy sensitivity or unexpected performance patterns observed across datasets:

\begin{itemize}[leftmargin=*, itemsep=0em]
    \item \textbf{Gemini 2.5 Flash on ReGrading Dataset (2JC) ($\sigma_{m,d} = 0.153$):} Under purely inductive reasoning, the model shows a strongly negative correlation (QWK = -0.20), whereas deductive reasoning yields a positive correlation of 0.26, a performance upgrade of 0.46. This suggests that example-based reasoning fundamentally misaligns with this particular dataset's grading criteria without a deductive structure.

    \item \textbf{LLaMA-4-Scout on IELTS Task 1 ($\sigma_{m,d} = 0.132$):} Ind+Ded achieves QWK $= 0.64$ while Ind+Abd achieves only $0.27$, a 2.4$\times$ difference. This indicates extreme sensitivity to which secondary reasoning mode accompanies inductive reasoning on descriptive writing tasks.

    \item \textbf{Gemini-2.5-Flash on Rice\_Chem} ($\sigma_{m,d} = 0.149$): driven by strong inductive performance (QWK = 0.70) but weak deductive performance (QWK = 0.51), suggesting chemistry content evaluation benefits strongly from example-based grounding for this model.

    \item \textbf{Gemini-2.5-Flash on CSEE ($\sigma_{m,d} = 0.109$):} Ded+Abd achieves only QWK $= 0.33$ while Ind+Abd and Ind+Ded both achieve $0.64$.  It suggests that inductive examples are critical for evaluating structured short-form writing tasks such as formal letters and compositions.

    \item \textbf{LLaMA-4-Scout on ASAP++ ($\sigma_{m,d} = 0.008$):} This dataset exhibits the lowest variance across models, suggesting that reasoning strategy has minimal impact. The overall task difficulty appears to dominate, with all strategies performing similarly poorly.

    \item \textbf{GPT-4o-mini on ASAP-AES ($\sigma_{m,d} = 0.018$):} Remarkably stable across all strategies (range: 0.915--0.954), suggesting that GPT-4o-mini evaluates essays consistently regardless of the specific reasoning configuration used.
\end{itemize}

\section{Exemplar Generalization}
\label{appendix:generalization}

Table \ref{tab:exemplar-gen} in Section~\ref{sec:results} reports exemplar generalization results under inductive 5-shot prompting where training examples are drawn from a different dataset than the test set. Here we provide additional context on the experimental design and observed failure modes not discussed in the main paper.

Dataset pairs were selected based on task compatibility: ASAP2$\rightarrow$ASAP-AES for within-domain AES transfer, and CSEE$\leftrightarrow$ASAP-SAS for cross-domain ASAG transfer. A single random seed (42) was used for exemplar selection, meaning results reflect one specific sampling, variability across seeds is characterized separately in Section~\ref{sec:exemplar_stability}.

The most notable failure is LLaMA-4-Scout under cross-paradigm transfer (ASAP-AES$\rightarrow$CSEE, $\Delta$QWK $= -95.5\%$; ASAP-SAS$\rightarrow$CSEE, $\Delta$QWK $= -47.2\%$), which is disproportionately large compared to GPT-4o-mini and Gemini-2.5-Flash on the same transfers. This collapse suggests LLaMA-4-Scout is particularly sensitive to domain mismatch between exemplars and test content, likely because its inductive reasoning relies more heavily on surface-level patterns in the provided examples rather than abstract quality criteria. In contrast, the unexpected improvement of Gemini-2.5-Flash under CSEE$\rightarrow$ASAP-AES transfer ($\Delta$QWK $= +12.5\%$) suggests that structured, form-oriented writing exemplars may positively scaffold student response evaluation for this model, an effect that warrants further investigation.

\definecolor{promptbg}{RGB}{245, 247, 250}
\definecolor{promptborder}{RGB}{70, 130, 180}
\definecolor{userbg}{RGB}{250, 245, 240}
\definecolor{userborder}{RGB}{180, 100, 70}
\definecolor{hybridbg}{RGB}{245, 250, 245}
\definecolor{hybridborder}{RGB}{70, 150, 70}

\tcbset{
    promptstyle/.style={
        colback=promptbg,
        colframe=promptborder,
        fonttitle=\small\bfseries,
        breakable,
        left=6pt, right=6pt, top=4pt, bottom=4pt,
        before upper={\setlength{\parindent}{0pt}\setlength{\parskip}{2pt}},
    },
    userstyle/.style={
        colback=userbg,
        colframe=userborder,
        fonttitle=\small\bfseries,
        breakable,
        left=6pt, right=6pt, top=4pt, bottom=4pt,
        before upper={\setlength{\parindent}{0pt}\setlength{\parskip}{2pt}},
    },
    hybridstyle/.style={
        colback=hybridbg,
        colframe=hybridborder,
        fonttitle=\small\bfseries,
        breakable,
        left=6pt, right=6pt, top=4pt, bottom=4pt,
        before upper={\setlength{\parindent}{0pt}\setlength{\parskip}{2pt}},
    },
}

\section{Prompt Templates}
\label{appendix:prompts}

Each reasoning strategy is implemented using a two-part prompt structure. The \textit{system prompt} specifies how the model should reason about the task, while the \textit{user prompt} provides the essay and clearly defines the expected output format. We describe the numeric essay-scoring variant here, as it represents the most general case. The classification variants follow the same structure, with the output restricted to one of three labels: \texttt{correct}, \texttt{incorrect}, or \texttt{contradictory}, depending on the task. For all strategies, we dynamically insert few-shot examples drawn from the training split of the corresponding dataset to guide the model's responses.
\subsection{Inductive Reasoning}

Inductive reasoning learns scoring patterns directly from training examples, inferring criteria from observed score-response pairs.

\begin{tcolorbox}[promptstyle, title={\textbf{System Prompt: Inductive}}]
{\small\ttfamily
You are an expert essay scorer using INDUCTIVE REASONING.\\[4pt]
INDUCTIVE PROCESS:\\
1. Learn scoring patterns from the examples below\\
2. Identify scoring criteria from the example patterns\\
3. Apply these learned patterns to score the new essay\\[4pt]
SCORING EXAMPLES FROM TRAINING DATA:\\
{[FEW-SHOT EXAMPLES INSERTED HERE]}\\[4pt]
From these examples, identify patterns in:\\
- What makes a high score vs low score\\
- How content quality affects scoring\\
- What level of development is expected\\[4pt]
SCORING RANGE: {[min, max]}\\
TASK: Score the essay based on the patterns you learned.
}
\end{tcolorbox}

\begin{tcolorbox}[userstyle, title={\textbf{User Prompt: Inductive}}]
{\small\ttfamily
Based on the patterns you learned from the examples above, score this essay:\\[4pt]
QUESTION/PROMPT:\\
{[QUESTION]}\\[4pt]
ESSAY TO SCORE:\\
{[STUDENT RESPONSE]}\\[4pt]
STOP. Do not write steps. Do not write explanations. Do not write reasoning.\\[4pt]
Your ENTIRE response must be EXACTLY one number between {[min]} and {[max]}.\\[4pt]
Nothing else. Just the number.
}
\end{tcolorbox}

\subsection{Deductive Reasoning}

Deductive reasoning applies general scoring rules and principles to the specific essay, deriving a score through logical rule application.

\begin{tcolorbox}[promptstyle, title={\textbf{System Prompt — Deductive}}]
{\small\ttfamily
You are an expert scorer using DEDUCTIVE REASONING.\\[4pt]
DEDUCTIVE PROCESS:\\
1. Start with GENERAL scoring rules/criteria\\
2. Apply rules to THIS SPECIFIC essay\\
3. Derive score logically from rule application\\[4pt]
EXAMPLES OF DEDUCTIVE SCORING:\\
GENERAL RULE 1: Complete answers provide required details\\
APPLICATION: Student provides 3 pieces of information\\
DEDUCTION: Meets requirement $\rightarrow$ Base score awarded\\[4pt]
GENERAL RULE 2: Answers must explain reasoning, not just state results\\
APPLICATION: Student gives correct number, no explanation\\
DEDUCTION: Partial credit only\\[4pt]
GENERAL RULE 3: Scientific answers must correctly apply relevant laws\\
APPLICATION: Student mentions law but misapplies concept\\
DEDUCTION: Partial credit (understands principle, has errors)\\[4pt]
SCORING RANGE: {[min, max]}\\
TASK: Essay scoring
}
\end{tcolorbox}

\begin{tcolorbox}[userstyle, title={\textbf{User Prompt: Deductive}}]
{\small\ttfamily
Apply deductive reasoning to score:\\[4pt]
QUESTION/PROMPT:\\
{[QUESTION]}\\[4pt]
ESSAY TO SCORE:\\
{[STUDENT RESPONSE]}\\[4pt]
Apply general scoring rules to this specific essay and derive the score.\\[4pt]
Provide ONLY a single number between {[min]} and {[max]} with no explanation. Just the number.
}
\end{tcolorbox}

\subsection{Abductive Reasoning}

Abductive reasoning treats the student response as an observation, generates possible explanations for the student's knowledge state, and infers the most plausible score from the best-fitting explanation. Although abductive reasoning is inherently inferential, the prompt requires the model to perform this reasoning internally while returning only the final numeric score, without exposing the reasoning traces.

\begin{tcolorbox}[promptstyle, title={\textbf{System Prompt: Abductive}}]
{\small\ttfamily
You are an expert scorer using ABDUCTIVE REASONING.\\[4pt]
ABDUCTIVE PROCESS:\\
1. Observe what the student wrote\\
2. Generate explanations for their knowledge state\\
3. Identify which explanation best fits the evidence\\
4. Infer appropriate score from that explanation\\[4pt]
EXAMPLES OF ABDUCTIVE SCORING:\\
OBSERVATION: Student provides amount, temperature, duration\\
POSSIBLE EXPLANATIONS:\\
- Student randomly listed items $\rightarrow$ irrelevant details expected\\
- Student identified genuinely missing info $\rightarrow$ important details expected\\
- Student did not read procedure $\rightarrow$ already-specified items expected\\
BEST EXPLANATION: Correctly identified critical missing details\\
SCORE: 2/3\\[4pt]
OBSERVATION: Student wrote correct number, no explanation\\
POSSIBLE EXPLANATIONS:\\
- Student ran simulation but does not understand why\\
- Student guessed correctly\\
- Student understands deeply\\
BEST EXPLANATION: Obtained correct result, cannot explain process\\
SCORE: 8/15\\[4pt]
SCORING RANGE: {[min, max]}\\
TASK: Essay scoring
}
\end{tcolorbox}

\begin{tcolorbox}[userstyle, title={\textbf{User Prompt: Abductive}}]
{\small\ttfamily
Use abductive reasoning to score:\\[4pt]
QUESTION/PROMPT:\\
{[QUESTION]}\\[4pt]
ESSAY (OBSERVATION):\\
{[STUDENT RESPONSE]}\\[4pt]
STOP. Do not write steps. Do not write explanations. Do not write reasoning.\\[4pt]
Your ENTIRE response must be EXACTLY one number between {[min]} and {[max]}.\\[4pt]
Nothing else. Just the number.
}
\end{tcolorbox}

\subsection{Hybrid Strategies}

The three hybrid strategies combine two reasoning phases sequentially within a single prompt. Each hybrid retains the full system prompt structure of both constituent strategies, with a combined instruction header directing the model to apply both reasoning processes before producing a score. Table~\ref{tab:hybrid_prompts} summarizes the process.

\begin{table}[H]
\centering
\resizebox{\columnwidth}{!}{
\begin{tabular}{lll}
\toprule
\textbf{Strategy} & \textbf{Phase 1} & \textbf{Phase 2} \\
\midrule
Ind+Ded & Learn patterns from examples (Inductive) & Apply general scoring rules (Deductive) \\
Ind+Abd & Learn patterns from examples (Inductive) & Infer best explanation (Abductive) \\
Ded+Abd & Apply general scoring rules (Deductive) & Infer best explanation (Abductive) \\
\bottomrule
\end{tabular}
}
\caption{Hybrid strategy prompt composition. Each hybrid combines the system prompt blocks of two solo strategies sequentially.}
\label{tab:hybrid_prompts}
\end{table}

\begin{tcolorbox}[hybridstyle, title={\textbf{System Prompt Structure: All Hybrid Strategies}}]
{\small\ttfamily
You are an expert scorer using [STRATEGY A] then [STRATEGY B] REASONING.\\[4pt]
PHASE 1 — [STRATEGY A] REASONING:\\
{[Full Phase 1 block from Strategy A]}\\[4pt]
PHASE 2 — [STRATEGY B] REASONING:\\
{[Full Phase 2 block from Strategy B]}\\[4pt]
COMBINED APPROACH:\\
1. Apply Strategy A reasoning\\
2. Apply Strategy B reasoning\\
3. Score based on both methods\\[4pt]
SCORING RANGE: {[min, max]}\\
Use both reasoning approaches before scoring.
}
\end{tcolorbox}

The user prompt for all hybrid strategies is identical to the solo strategies, constraining output to a single number within the valid scoring range. Full prompt code for all six strategies is available in our repository.\footnote{\url{https://github.com/nlpatunt/sgrades-experiments}}

\section{Prediction Stability Analysis}
\label{app:stability}

This appendix presents additional analysis of prediction stability across repeated model evaluations. The goal of this experiment is to measure how consistently LLMs produce the same predictions when evaluated multiple times on identical inputs. This analysis is distinct from the exemplar-sampling stability experiment reported in Section~\ref{sec:results}, which varies few-shot exemplar seeds; here, we measure run-to-run prediction variability under repeated querying of the same response.

\subsection{Experimental Protocol}
For each model--strategy combination, prediction stability was measured by generating \textbf{three separate predictions} for every student response.

To make repeated evaluation computationally feasible across models and reasoning strategies, stability analysis was conducted on a sampled subset comprising \textbf{21\% of each dataset's test split}. The same sampled subset was reused across all models and strategies.

All runs used a \textbf{sampling temperature of 0.1}, consistent with the settings used in the primary experiments, which keeps model outputs nearly deterministic. As a result, any variation observed across repeated runs primarily reflects inconsistencies in the model's scoring behavior rather than sampling-induced randomness.

For strategies that include an inductive component (Ind, Ind–Ded, Ind–Abd), few-shot examples are drawn from the training split using \textbf{random seed 42}, ensuring prompt construction and example selection remain identical across repeated runs.

\subsection{Stability Metrics}

We evaluate stability separately for numeric and categorical datasets.

\paragraph{Numeric datasets.}
For each essay $i$, let $\hat{y}_i^{(1)}, \hat{y}_i^{(2)}, \hat{y}_i^{(3)}$ be the scores produced by the three inference calls. We compute the sample standard deviation:

\begin{equation}
\sigma_i = \sqrt{\frac{1}{2}\sum_{k=1}^{3}\left(\hat{y}_i^{(k)} - \bar{y}_i\right)^2},
\qquad
\bar{y}_i = \frac{1}{3}\sum_{k=1}^{3}\hat{y}_i^{(k)}.
\end{equation}

We report the \textbf{pooled mean standard deviation} $\overline{\sigma}$
across all essays, weighted by essay count. Sub-question datasets
(e.g., OS Dataset and Rice Chem) are first merged to avoid double-counting.
\textbf{Lower values indicate more stable scoring.}

\paragraph{Categorical datasets.}
For categorical tasks, stability is measured using agreement across the three predictions. For essay $i$:

\begin{equation}
a_i =
\frac{\left|\{k : \hat{y}_i^{(k)} =
\mathrm{mode}(\hat{y}_i^{(1)},\hat{y}_i^{(2)},\hat{y}_i^{(3)})\}\right|}{3}.
\end{equation}

The final predicted label is taken as the \textbf{mode} of the three predictions. For 2-way tasks, this corresponds to choosing between
\textit{correct} and \textit{incorrect}; for 3-way tasks, it corresponds to
choosing among \textit{correct}, \textit{incorrect}, and \textit{contradictory}.
If all three predictions differ, the response is \textbf{not discarded}; its agreement score is $1/3$, reflecting maximal disagreement across the three calls.

We report the \textbf{pooled mean agreement} across all categorical datasets. Higher values indicate more stable predictions.

\subsection{Results and Discussion}

\subsubsection{Overall Stability Across Models and Strategies}

\begin{figure}[!htbp]
    \centering
    \includegraphics[width=0.85\linewidth]{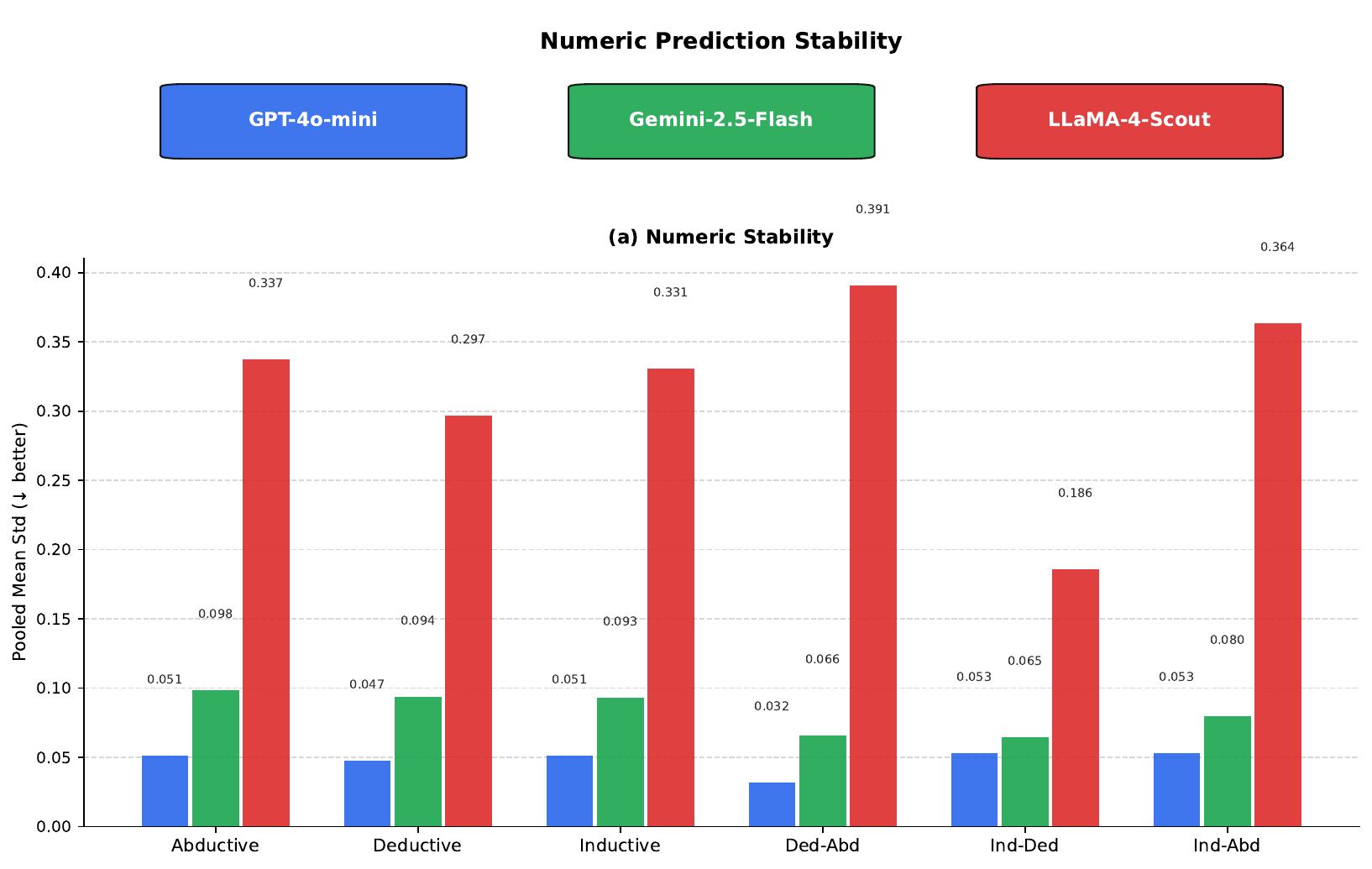}
    \caption{Pooled mean standard deviation ($\overline{\sigma}$) on numeric
    datasets across six reasoning strategies and three models. Lower is better.}
    \label{fig:numeric_stability}
\end{figure}

Figure~\ref{fig:numeric_stability} summarizes stability for numeric scoring datasets using pooled mean standard deviation across three repeated calls. Clear differences emerge across the three models. GPT-4o-mini is the most stable overall, with pooled mean standard deviation values ranging from approximately 0.032 to 0.053 across reasoning strategies. Gemini-2.5-Flash shows slightly higher but still relatively low variability, with values ranging from about 0.065 to 0.098. In contrast, LLaMA-4-Scout exhibits substantially higher instability, with pooled mean standard deviation values between 0.186 and 0.391 across strategies. 

Among the six strategies, \textit{Ded+Abd} yields the lowest numeric instability for GPT-4o-mini, while \textit{Ind+Ded} gives the lowest value for both Gemini-2.5-Flash and LLaMA-4-Scout. Overall, the gap between models is much larger than the variation introduced by strategy choice alone, indicating that model choice plays the dominant role in the stability of numeric predictions. 

\begin{figure}[!htbp]
    \centering
    \includegraphics[width=0.85\linewidth]{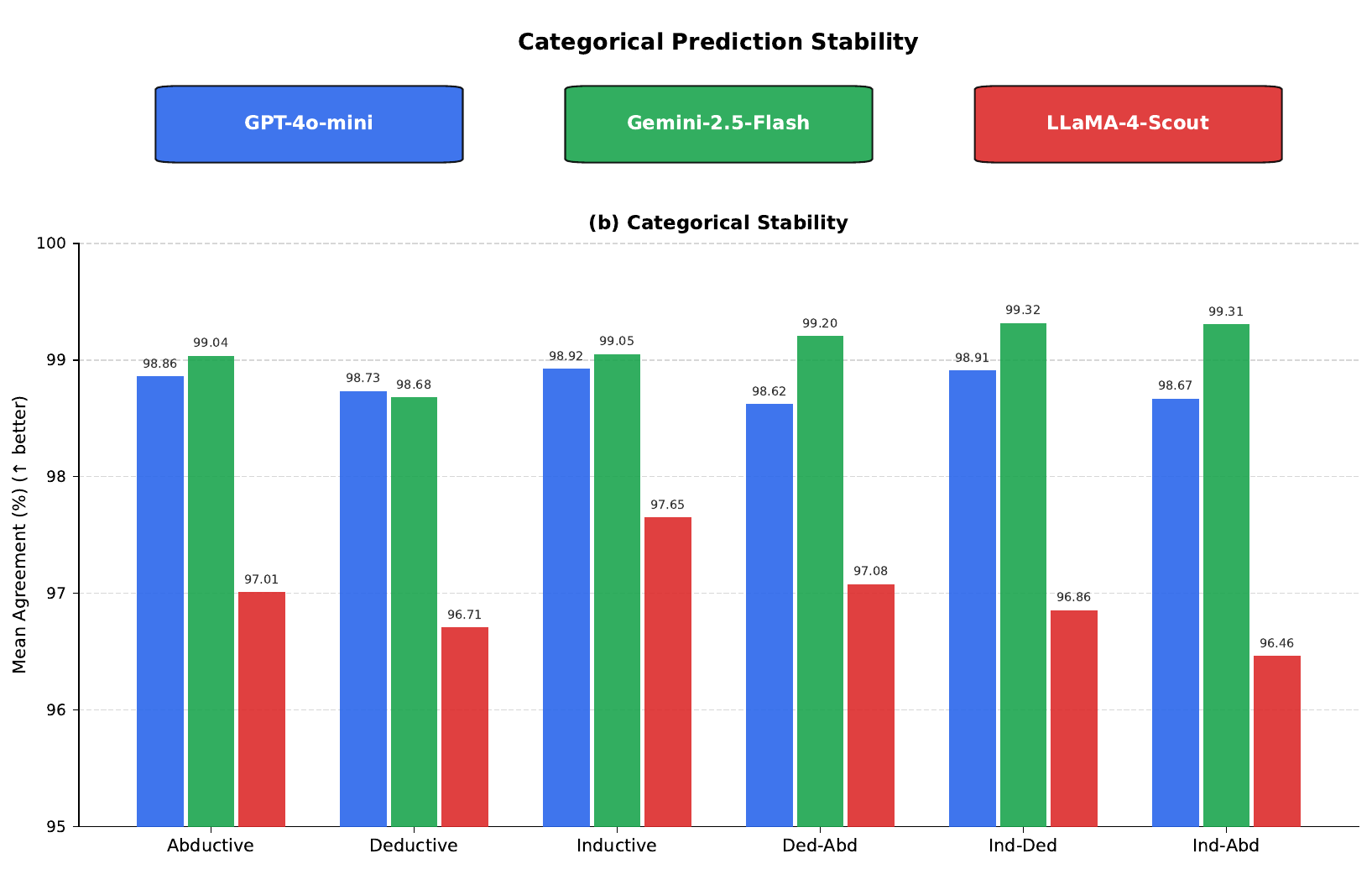}
    \caption{Pooled mean agreement (\%) on categorical datasets across six
    reasoning strategies and three models. Higher is better.}
    \label{fig:categorical_stability}
\end{figure}

Figure~\ref{fig:categorical_stability} presents the corresponding results for categorical datasets using mean agreement across repeated calls. All three models are considerably more stable on categorical tasks than on numeric scoring tasks. \textbf{GPT-4o-mini and Gemini-2.5-Flash} achieve near-perfect agreement across all reasoning strategies, generally remaining between 98.6\% and 99.3\%. LLaMA-4-Scout is less stable than the other two models, but still maintains relatively high agreement, ranging from about 96.5\% to 97.7\%. These results suggest that categorical grading is inherently less sensitive to variability in repeated inference than numeric scoring, likely because classification decisions involve a smaller output space and fewer fine-grained distinctions than essay score prediction.

Taken together, these findings show that prediction stability depends primarily on model choice and task type rather than reasoning strategy. Numeric scoring tasks are substantially more variable than categorical tasks across all models. GPT-4o-mini is the most stable on numeric tasks, Gemini-2.5-Flash leads on categorical tasks, and LLaMA-4-Scout is consistently the least stable under repeated evaluation.

\subsubsection{Dataset-Level Stability Patterns}
\begin{figure}[!htbp]
    \centering
    \includegraphics[width=0.85\linewidth]{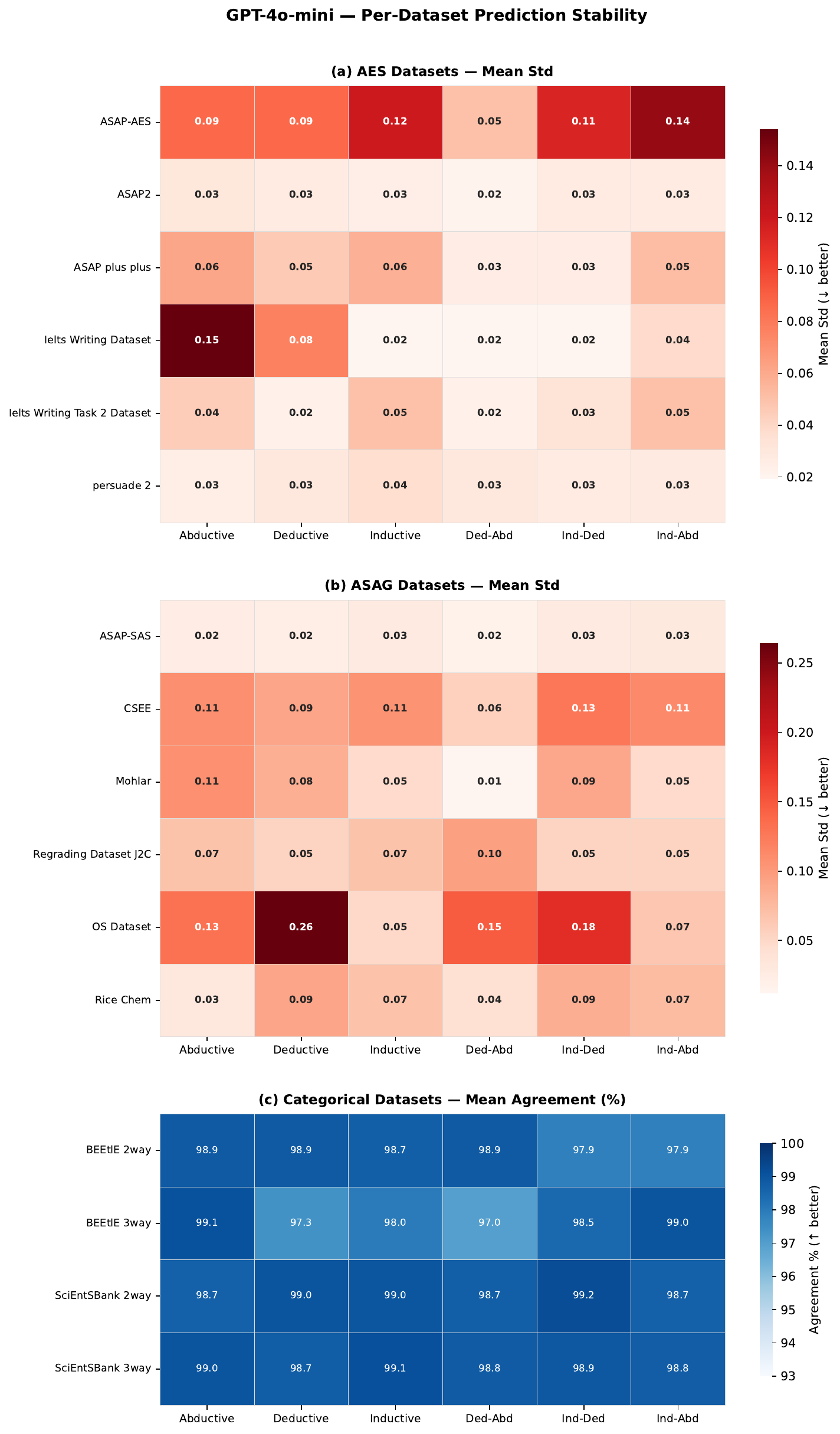}
    \caption{Per-dataset prediction stability for GPT-4o-mini across six
    reasoning strategies.}
    \label{fig:gpt_heatmap}
\end{figure}

\begin{figure}[!htbp]
    \centering
    \includegraphics[width=0.85\linewidth]{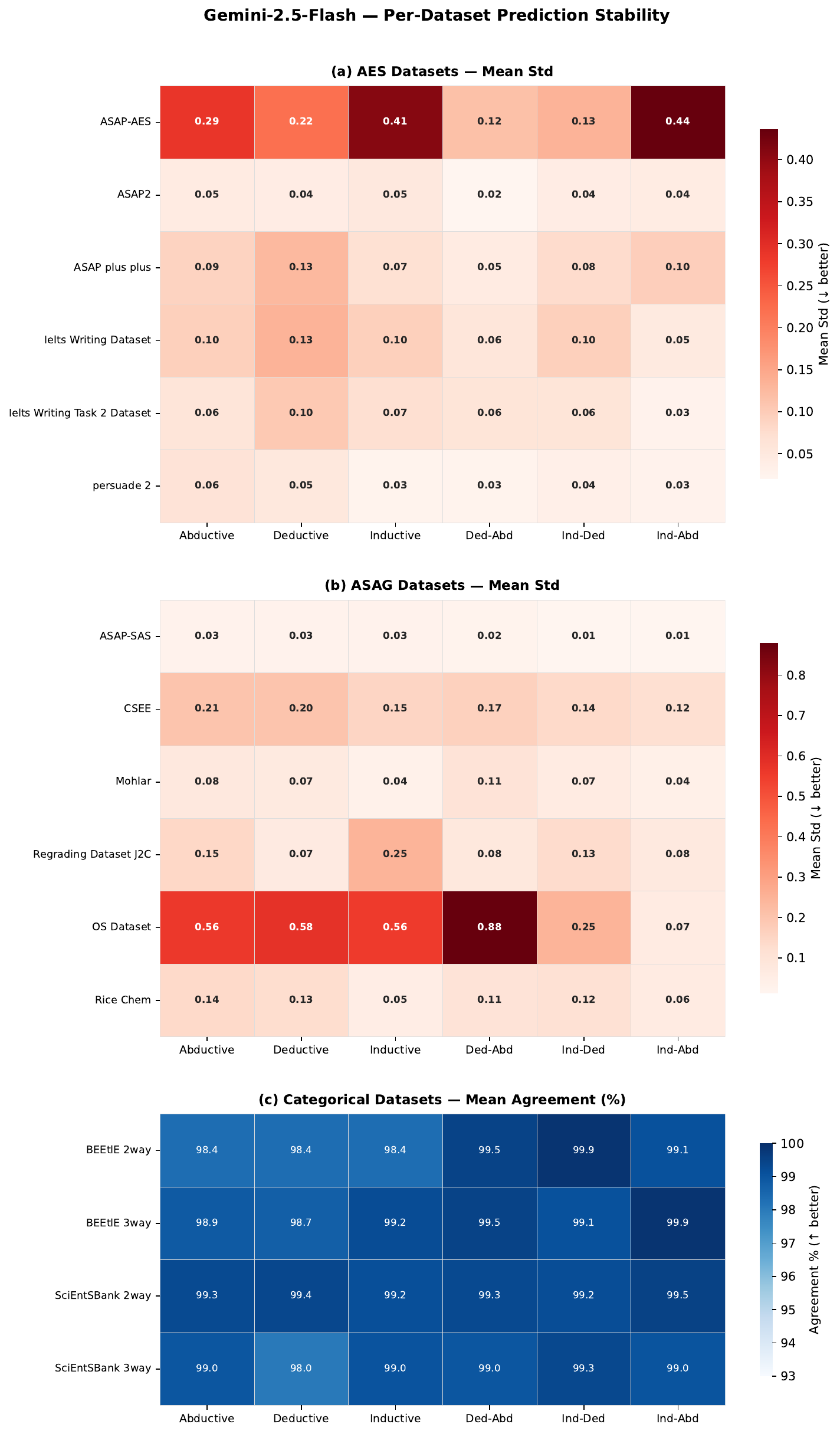}
    \caption{Per-dataset prediction stability for Gemini-2.5-Flash across six reasoning strategies.}
    \label{fig:gemini_heatmap}
\end{figure}

\begin{figure}[!htbp]
    \centering
    \includegraphics[width=0.85\linewidth]{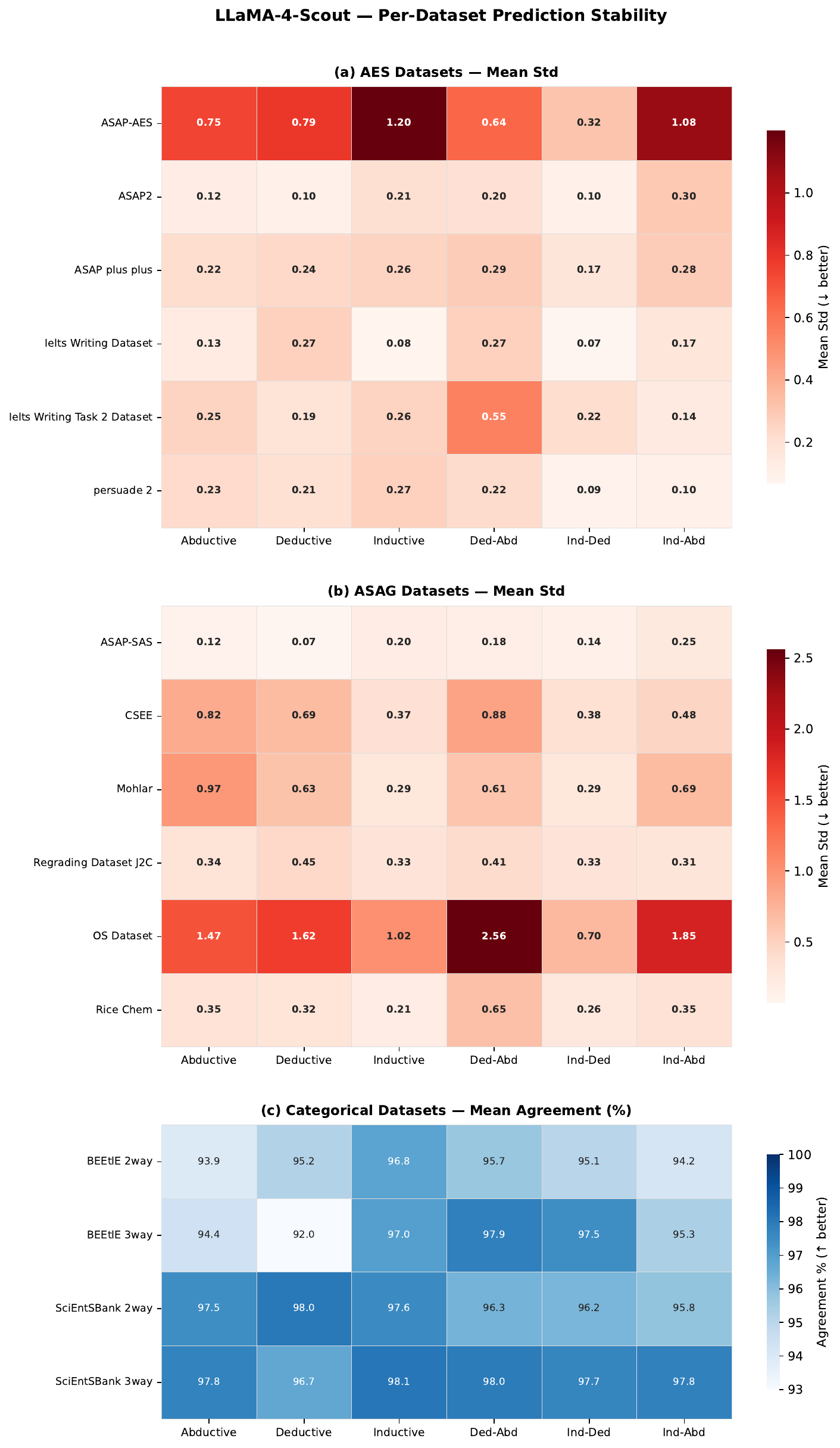}
    \caption{Per-dataset prediction stability for LLaMA-4-Scout across six reasoning strategies.}
    \label{fig:llama_heatmap}
\end{figure}

Figures~\ref{fig:gpt_heatmap}, \ref{fig:gemini_heatmap}, and \ref{fig:llama_heatmap} present per-dataset stability heatmaps for all three models. A clear pattern is that \textbf{ASAG predictions are consistently less stable than AES predictions} across all models. 

Averaged across all datasets and strategies, ASAG mean standard deviation is 1.6$\times$ higher than AES for GPT-4o-mini (0.077 vs.\ 0.049) and Gemini-2.5-Flash (0.160 vs.\ 0.098), and 2.0$\times$ higher for LLaMA-4-Scout (0.599 vs.\ 0.305). This gap persists even after excluding OS\_Dataset --- which is an outlier due to its extremely small test split (3--8 essays per sub-question) --- for GPT-4o-mini and LLaMA-4-Scout (1.3$\times$ in both cases), suggesting that the difference reflects broader task-level characteristics rather than only a single dataset artifact.

Within ASAG, \textbf{CSEE} and \textbf{Mohlar} are consistently the most
unstable datasets after OS\_Dataset, while \textbf{ASAP-SAS} and
\textbf{Rice\_Chem} remain stable across all models. Within AES, \textbf{ASAP-AES} shows the highest variability across models.

Overall, these heatmaps reinforce the broader conclusion that prediction stability is shaped by both model and task type. While the prompting strategy has some effect, the larger and more consistent difference appears between AES and ASAG, with short-answer grading showing substantially greater sensitivity to repeated evaluation. This supports the view that stability should be considered a task-dependent property of LLM-based scoring systems rather than a uniform model behavior.

\end{document}